\documentclass{article}

\PassOptionsToPackage{numbers, compress, sort}{natbib}

\usepackage[preprint]{neurips_data_2022}

\usepackage[utf8]{inputenc} %
\usepackage[T1]{fontenc}    %
\usepackage{hyperref}       %
\usepackage{url}            %
\usepackage{booktabs}       %
\usepackage{amsfonts}       %
\usepackage{nicefrac}       %
\usepackage{microtype}      %
\usepackage{xcolor}         %
\usepackage{natbib}
\usepackage{epigraph}
\usepackage{amssymb}%
\usepackage{pifont}%
\usepackage{graphicx}
\usepackage{caption}
\usepackage{subcaption}
\usepackage{listings}
\usepackage{xcolor}

\colorlet{punct}{red!60!black}
\definecolor{background}{HTML}{EEEEEE}
\definecolor{delim}{RGB}{20,105,176}
\colorlet{numb}{magenta!60!black}

\lstdefinelanguage{json}{
    basicstyle=\small\ttfamily,
    numbers=none,
    numberstyle=\scriptsize,
    stepnumber=1,
    numbersep=8pt,
    showstringspaces=false,
    breaklines=true,
    frame=lines,
    literate=
     *{0}{{{\color{numb}0}}}{1}
      {1}{{{\color{numb}1}}}{1}
      {2}{{{\color{numb}2}}}{1}
      {3}{{{\color{numb}3}}}{1}
      {4}{{{\color{numb}4}}}{1}
      {5}{{{\color{numb}5}}}{1}
      {6}{{{\color{numb}6}}}{1}
      {7}{{{\color{numb}7}}}{1}
      {8}{{{\color{numb}8}}}{1}
      {9}{{{\color{numb}9}}}{1}
      {:}{{{\color{punct}{:}}}}{1}
      {,}{{{\color{punct}{,}}}}{1}
      {\{}{{{\color{delim}{\{}}}}{1}
      {\}}{{{\color{delim}{\}}}}}{1}
      {[}{{{\color{delim}{[}}}}{1}
      {]}{{{\color{delim}{]}}}}{1},
}

\newbox{\bigpicturebox}
\usepackage{algorithm}
\usepackage{algorithmic}
\usepackage{xspace}
\usepackage{bm}

\title{The ArtBench Dataset: Benchmarking Generative Models with Artworks}

\author{%
  Peiyuan Liao\thanks{Equal contribution.} \\
  Carnegie Mellon University\\
  Pittsburgh, PA 15213 \\
  \texttt{peiyuanl@cs.cmu.edu} \\
  \And Xiuyu Li$^*$ \\
  University of California, Berkeley\\
  Berkeley, CA 94720 \\
  \texttt{xiuyu@berkeley.edu} \\
  \And Xihui Liu \\
  University of California, Berkeley\\
  Berkeley, CA 94720 \\
  \texttt{xihui.liu@berkeley.edu} \\
  \And Kurt Keutzer \\
  University of California, Berkeley\\
  Berkeley, CA 94720 \\
  \texttt{keutzer@berkeley.edu}
}

\begin{document}

\maketitle

\newcommand{\sect}[1]{Section~\ref{#1}}
\newcommand{\ssect}[1]{\S~\ref{#1}}
\newcommand{\eqn}[1]{Equation~\ref{#1}}
\newcommand{\fig}[1]{Figure~\ref{#1}}
\newcommand{\tbl}[1]{Table~\ref{#1}}
\newcommand{\algo}[1]{Algorithm~\ref{#1}}
\newcommand{\name}{ArtBench-10\xspace}
\newcommand{\myparagraph}[1]{\noindent \textbf{#1}}
\newcommand{\na}{---}

\newcommand{\peiyuan}[1]{\textcolor{blue}{[Peiyuan: #1]}}
\newcommand{\xiuyu}[1]{\textcolor{violet}{[Xiuyu: #1]}}

\begin{abstract}
We introduce ArtBench-10, the first class-balanced, high-quality, cleanly annotated, and standardized dataset for benchmarking artwork generation. It comprises 60,000 images of artwork from 10 distinctive artistic styles, with 5,000 training images and 1,000 testing images per style. ArtBench-10 has several advantages over previous artwork datasets. Firstly, it is \textit{class-balanced} while most previous artwork datasets suffer from the long tail class distributions. Secondly, the images are of \textit{high quality} with \textit{clean annotations}. Thirdly, ArtBench-10 is created with \textit{standardized} data collection, annotation, filtering, and preprocessing procedures. We provide three versions of the dataset with different resolutions ($32\times32$, $256\times256$, and original image size), formatted in a way that is easy to be incorporated by popular machine learning frameworks. We also conduct extensive benchmarking experiments using representative image synthesis models with ArtBench-10 and present in-depth analysis. The dataset is available at \url{https://github.com/liaopeiyuan/artbench} under a Fair Use license.
\end{abstract}

\section{Introduction}
\epigraph{A painting is not a picture of an experience, but is the experience.}{\textit{Mark Rothko}}

\begin{figure}[h]
\centering
\includegraphics[trim={1cm 0.5cm 1cm 0},clip, width=\linewidth]{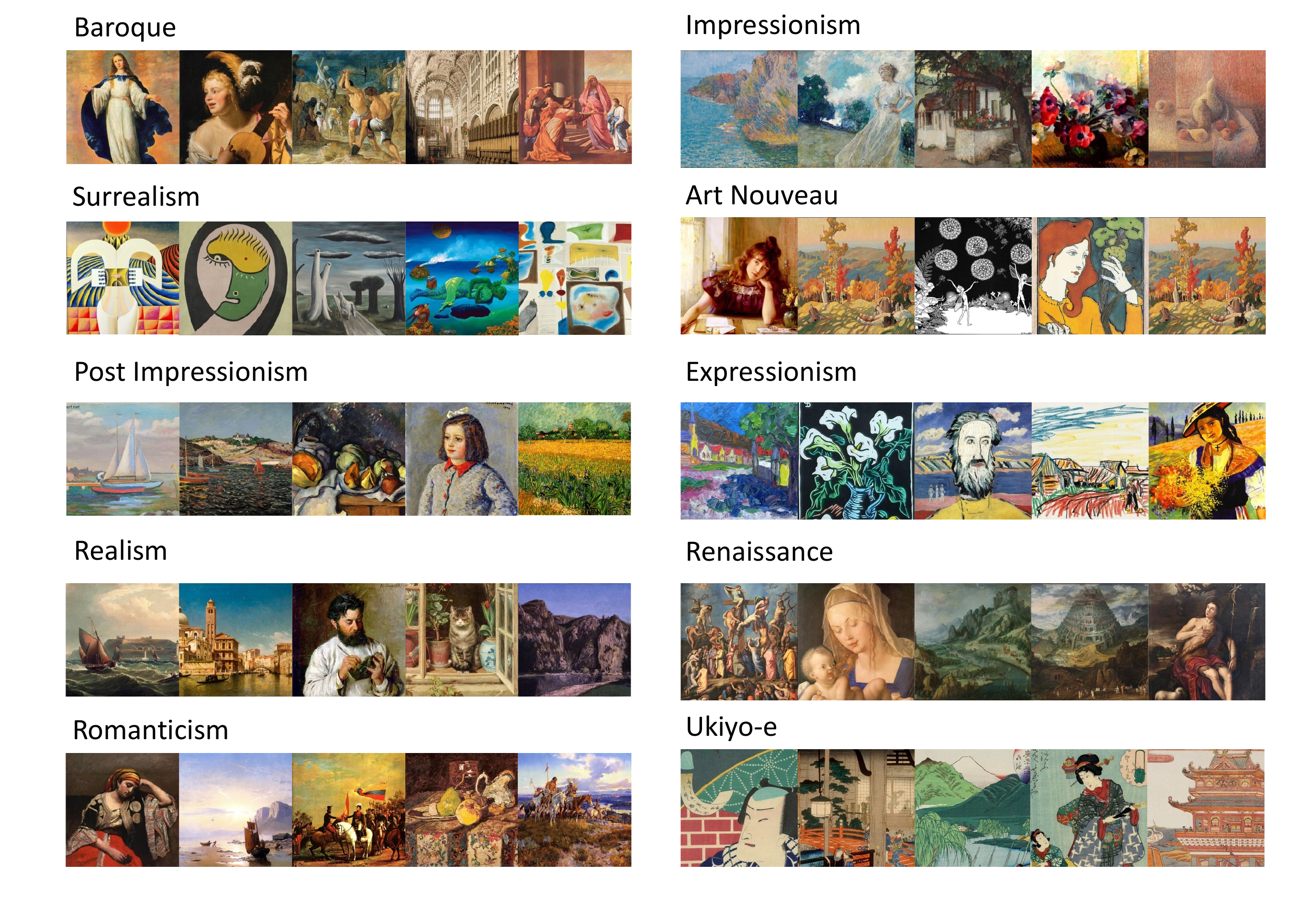}
\caption{Overview of the 10 artistic styles and corresponding images in ArtBench-10. ArtBench-10 is a class-balanced dataset with 6,000 images for each of the 10 artistic styles.}
 \label{overview}
 \end{figure}

Deep generative models \cite{brock2018large, sauer2022stylegan, karras2020analyzing, ho2020denoising, rombach2021highresolution, dhariwal2021diffusion} are capable of modeling complex, high dimensional distributions, and one of the prominent applications is the synthesizing diverse and high-fidelity images. 
Through extensive training on curated datasets, recent works are able to generate images with highly coherent semantic details and global composition.
In the meantime, computational understanding of art has been the interest of many researchers and creators for its implication to art history, computational creativity and human-computer interaction \cite{lang2021transforming, mccormack2001art, 10.2307/45149256}.
In particular, generating artworks with various styles has attracted more and more attention.

While there have been various standardized image synthesis benchmarks such as CIFAR-10~\cite{krizhevsky2009learning}, ImageNet~\cite{deng2009imagenet}, and LSUN~\cite{yu2015lsun}, they are biased towards photos of real scenes or objects rather than artwork.
Recent works like DALLE 2~\cite{ramesh2022hierarchical}, GLIDE~\cite{nichol2021glide}, and Imagen~\cite{saharia2022photorealistic} trained on large-scale datasets have demonstrated the potential of generative models to generate artworks.
However, those models are trained on a mixture of photos and artworks of various styles, making it difficult to conduct analysis and evaluations on artwork generation of particular artistic styles.

On the other hand, there were previous efforts to collect artwork datasets mainly for artwork classification and artwork attribute prediction.
The previous artwork datasets contain a wide range of digital collections of artworks ranging from images of exhibits in the Metropolitan Art Museum \cite{the_met_2000} to 3D scans of Mogao Caves \cite{dunhuang_academy_2015}. The artworks are annotated with attributes, emotions, descriptions, or instances.
However, those datasets are not suitable for benchmarking artwork synthesis methods because of the long tail class distributions.
Moreover, there were no standard data collection, annotation, filtering, and preprocessing processes for creation of those datasets, leading to a series of problems such as low-quality images, noisy labels, and duplicate artworks in the dataset. Thus, a standardized, balanced, high-quality, and clean benchmark for artworks generation is desired.

In this work, we propose ArtBench-10, the first class-balanced, high-quality, cleanly-annotated, and standardized benchmark for artworks synthesis.
The dataset is composed of 60,000 images annotated with 10 artistic styles, as illustrated in~\fig{overview}.
We carefully design the standardized data collection, annotation, filtering, and preprocessing schemes to ensure that the images are of high-quality with clean and balanced labels.
We provide three versions of the dataset with $32\times32$, $256\times256$, and original resolutions, respectively, with standard formats that are easy to be incorporated with existing dataloaders of commonly used deep learning frameworks such as PyTorch and Tensorflow.
We conduct dataset statistics analysis to demonstrate the advantages of our dataset over previous artworks datasets.
We also benchmark several representative image synthesis approaches on ArtBench-10 and provide comprehensive qualitative and quantitative analysis on those approaches.

\section{Related Work}\label{sec:related}
\paragraph{Image Synthesis}
Image synthesis aims to generate synthetic images that follow the distribution of real images. 
Deep generative models have made great progress in image synthesis in the recent years.
Among them the most popular types of generative models are variational auto-encoder (VAE)~\cite{kingma2014auto}, flow-based models, generative adversarial networks (GAN)~\cite{goodfellow2014generative}, and diffusion-based models~\cite{sohl2015deep, song2019generative, ho2020denoising}. 
VAE is a stochastic variational inference and learning algorithm for modeling the data distribution.
Flow-based deep generative models introduce normalizing flows, a powerful statistics tool for density estimation.
GANs are composed of generator and discriminator that are trained in an adversarial way: the discriminator aims to discriminate between fake and real images, and the generator aims to generate fake images that can fool the discriminator.
Diffusion models formulate a Markov process to iteratively add random noise to images and then learn the reverse process to synthesis images by gradually removing noise from the sampled random noise.
GANs have dominated the field of image synthesis for years~\cite{karras2019style,karras2020analyzing,karras2020training,karras2021alias,suer2021projected,brock2018large} before recently some diffusion models~\cite{ho2020denoising,liu2022pseudo,rombach2021highresolution,nichol2021improved} are shown to perform on par or even better than state-of-the-art GANs with more stable training procedures.

Conditional image synthesis aims to generate synthetic images according to the given conditional information. 
Among different types of condition information, class label is the most common one. 
Our proposed ArtBench-10 is annotated with 10 artistic styles, so it can be used for benchmarking both unconditional and class-conditional artwork synthesis.

\begin{table}[t]
  \caption{A selection of existing datasets for benchmarking generative models, compared to ArtBench. LSUN datasets are usually trained and evaluated on a per-class basis, such as LSUN Car, LSUN Bedroom, etc., so each of them is considered as unconditional in this comparison.}
  \label{benchmarks}
  \vspace{0.5em}
  \centering
  \begin{tabular}{lllll}
    \toprule
    Name     & \# Classes     & Resolution & \# Images & Domain \\
    \midrule
    Pokemon \cite{sauer2022stylegan} & N/A & $256^2$, $1024^2$ & 834 & "Pokemons" \\
    MetFaces \cite{karras2020training} & N/A & $1024^2$ & 1336 & human faces from art \\
    STL-10 \cite{pmlr-v15-coates11a} &  10 & $96^2$ & 13,000 & daily objects \\
    MNIST \cite{mnist} &  10 & $28^2$ & 70,000 & handwritten digits \\
    FFHQ \cite{karras2019style} & N/A & $256^2$, $1024^2$  & 70,000 & faces from Flickr \\
    CelebA \cite{liu2015faceattributes} & N/A & $64^2$, $256^2$, $1024^2$ & 202,599 & celebrity faces      \\
    ImageNet \cite{deng2009imagenet} & 1000 & $32^2$, $64^2$, $128^2$, $256^2$ & 1,431,167 & objects in WordNet \\
    LSUN \cite{yu2015lsun} & N/A  & $256^2$, Varied & 120,000 to 3,000,000 & objects and scenes    \\
    \midrule
    ArtBench-10 & 10 & $32^2$, $256^2$, Varied & 60,000 & artworks \\
    \bottomrule
  \end{tabular}
\end{table}

\paragraph{Image Synthesis Datasets and Benchmarks}
There are many commonly used image synthesis benchmarks, as listed in \tbl{benchmarks}. 
Some datasets only contain images of a single category, used for benchmarking unconditional image synthesis. 
Representative datasets of this type are Pokemons from Pokemon dataset~\cite{sauer2022stylegan}, human face photos from FFHQ~\cite{karras2019style} and CelebA~\cite{liu2015faceattributes}, and artworks of faces from MetFaces~\cite{karras2020training}. 
LSUN~\cite{yu2015lsun} has multiple categories but it is usually used for unconditional image synthesis with only one category from the dataset.
Other datasets are multi-class datasets, including MNIST~\cite{mnist}, STL-10~\cite{pmlr-v15-coates11a}, and ImageNet~\cite{deng2009imagenet}, for benchmarking class-conditional image synthesis.
However, most of them are biased towards photos and contain few artworks.

\begin{table}[h]
  \centering
  \caption{Comparison of ArtBench-10 to other art datasets. Our ArtBench-10 is class-balanced and created with standardized data collection, annotation, filtering, and preprocessing pipelines. We include more details on data collection, cleaning, and preprocessing of previous datasets in supplementary materials \ref{apppend:prev_datasets}.}
  \label{exisiting_datasets}
  \vspace{0.5em}
  \resizebox{\textwidth}{!}{
  \begin{tabular}{llccllccll}
    \toprule
    Domain & Name  & \# Images & \# Classes & Type of annotations & Image source & Balanced & Standardized \\
    \midrule
    Paintings & VGG Paintings \cite{Crowley2014TheSO} & 8,629 & 10 & Object category & Art UK & \ding{55} & \ding{55}  \\  %
     & SemArt \cite{Garca2018HowTR} & 21,383 & 21,383 & Art attributes, descriptions & Web Gallery of Art & \ding{55} & \ding{51} \\ %
     & WikiPaintings \cite{Karayev2014RecognizingIS} & 85,000 & 25 & Style & WikiArt & \ding{55} & \ding{55} \\ %
    \midrule
    Prints & PrintArt \cite{Carneiro2012ArtisticIC} & 988 & 75 & Art theme & Artstor & \ding{55} & \ding{55} \\ %
    \midrule
    Digital art & BAM \cite{Wilber2017BAMTB} & 65M & 9 & Media, content, emotion & Enhance & \ding{55} & \ding{55} \\ %
    \midrule
    Artwork & Open MIC \cite{Koniusz2018MuseumEI} & 16,156 & 866 & Instance & \textit{Authors} & \ding{55} & \ding{55} \\ %
     & NoisyArt \cite{DelChiaro2019NoisyArtAD} & 89,095 & 3,120 & Instance (noisy) & \textit{Various} & \ding{55} & \ding{55} \\ %
     & Rijksmuseum \cite{Mensink2014TheRC} & 112,039 & 6,629 & Art attributes & Rijksmuseum & \ding{55} & \ding{51} \\ %
     & iMET \cite{Zhang2019TheIC} & 155,531 & 1,103 & Concepts & The Met & \ding{55} & \ding{55} \\ %
     & The Met \cite{ypsilantis2021met} & 418,605 & 224,408 & Instance & \textit{Various} & \ding{55} & \ding{51} \\ %
     & Art500k \cite{Mao2017DeepArtLJ} & 554,198 & 1,000 & Art attributes & \textit{Various} & \ding{55} & \ding{55} \\ %
     & OmniArt \cite{Strezoski2018OmniArtAL} & 1,348,017 & 100,433 & Art attributes & \textit{Various} & \ding{55} & \ding{51} \\ %
    \midrule
    Artwork & ArtBench-10 & 60,000 & 10 & Style & \textit{Various} & \ding{51} & \ding{51} \\
    \bottomrule
  \end{tabular}
}
\end{table}

\paragraph{Artworks Datasets and Benchmarks}
Most previous artworks datasets, as shown in \tbl{exisiting_datasets}\footnote{Part of the table is inherited from the table in~\citet{ypsilantis2021met}},  are targeted for artwork classification and artwork attribute prediction.
However, they are in general not suitable for benchmarking artwork synthesis, because of the long tail class distribution and undesirable artwork types for image synthesis.
Several previous works designed curated subsets of certain categories for artwork synthesis evaluation.
For instance, \citet{xue2021end} manually processed 2192 high-quality traditional Chinese landscape paintings from Metropolitan Museum of Art etc.
\citet{zhu2017unpaired} featured a customized dataset by hand pruning scraped data from WikiArt. 
\citet{tan2017artgan} scraped 80,000 annotated artworks from WikiArt to conduct conditional image generation experiments on ``genre'', ``artist'' and ``style'' classes respectively. %
Despite the curated selection and processing, they still suffer from several limitations such as imbalanced classes, inconsistent image quality, duplicate images, noisy labels, and non-standardized pre-processing.
Our work propose the first \textit{class-balanced, high-quality, cleanly-labeled, standardized} artwork synthesis benchmark, ArtBench-10.

\section{The ArtBench-10 Dataset}\label{sec:dset}

ArtBench-10 is a dataset of 60,000 artworks from 10 distinctive artistic styles across human history. 
The dataset comprises mainly of paintings between the 14th century and the 21th century, but also includes murals and sculptures from the same period of time.
The 10 artistic styles of our dataset and representative images from each style are illustrated in \fig{overview}.
The dataset is made available under a Fair Use license per the requirements from the image sources, while we also provide a public-domain subset that allows unlimited commercial use. This will be specified in the metadata schema in supplementary materials~\ref{apppend:metadata}.

\subsection{Limitation of Existing Artwork Datasets}

\begin{figure}[h]
\begin{subfigure}[h]{.59\linewidth}
  \centering
    \includegraphics[width=0.85\linewidth]{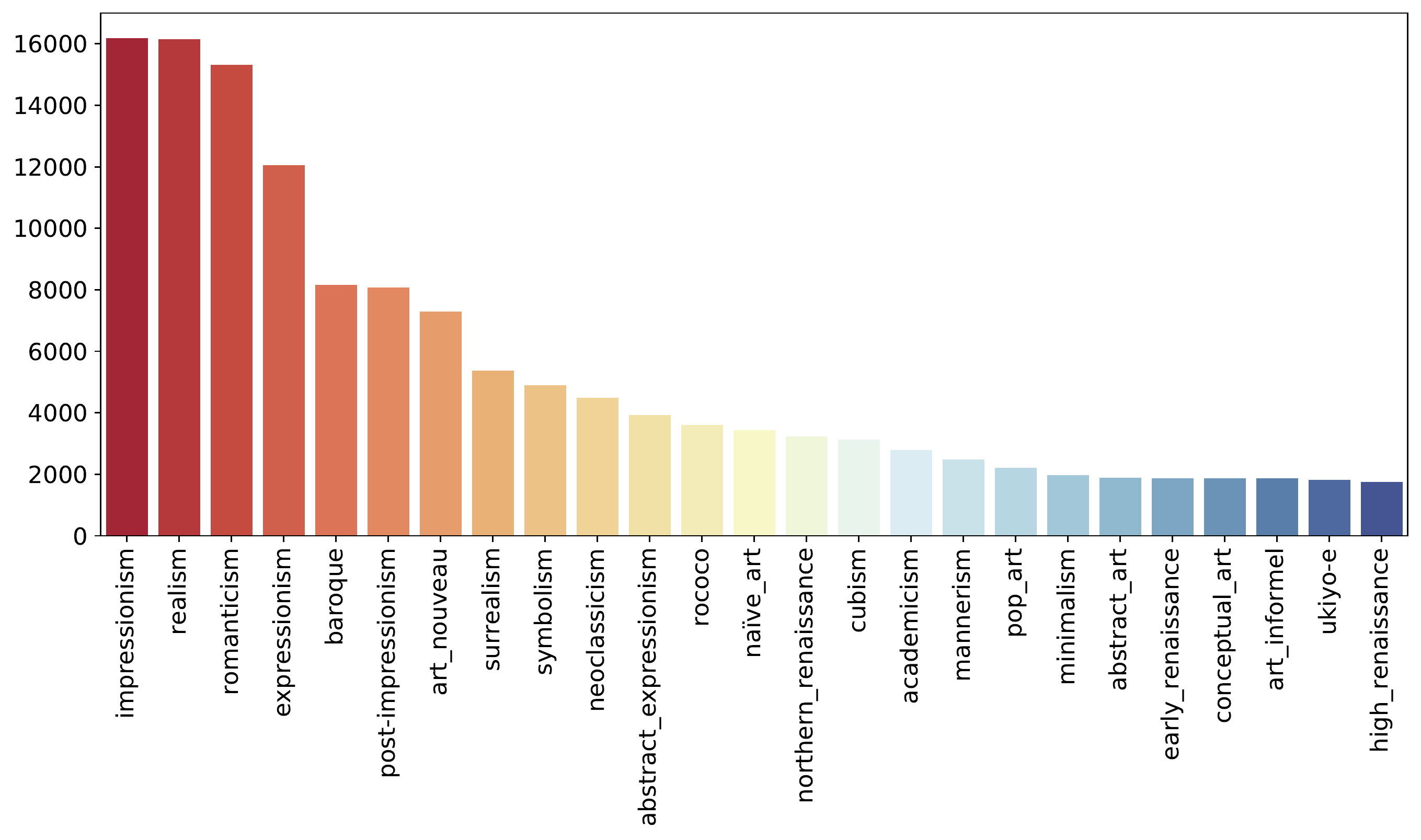}
  \label{fig:sub2}
\end{subfigure}
\begin{subfigure}[h]{.4\linewidth}
  \centering
    \includegraphics[width=\linewidth]{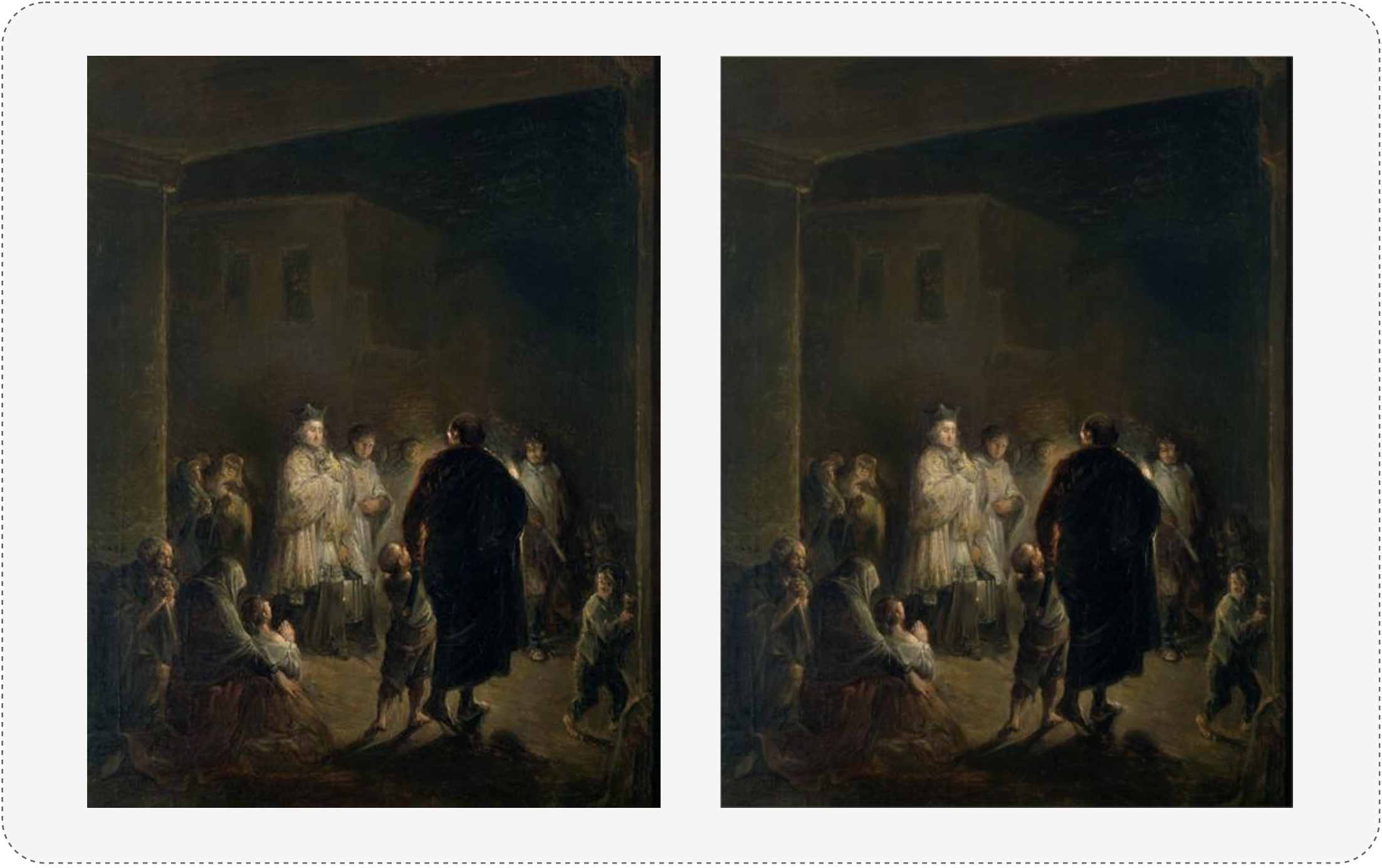}
  \label{fig:sub3}
\end{subfigure}
\caption[Caption for LOF]{Issues with the scraped WikiArt data (97908 images as in April 2022). (Left) Long tail distribution over classes. (Right) A pair of duplicate images in the dataset.\protect\footnotemark}
\label{wikiart_curated}
\end{figure}

There have been several previous artwork datasets in literature, as shown in \tbl{exisiting_datasets}. 
However, previous artwork datasets are mainly collected for visual recognition tasks, such as attribute prediction and artwork classification, and they are not suitable for benchmarking artwork synthesis.
Some datasets~\cite{Mensink2014TheRC, Koniusz2018MuseumEI} contain certain types of artworks (e.g. photos, porcelains, etc.) that are undesirable for artworks generation.
Some datasets~\cite{ypsilantis2021met, Garca2018HowTR, Strezoski2018OmniArtAL} are with instance-level annotations which can not be used for class-conditional artworks synthesis.
Previous works~\cite{xue2021end,zhu2017unpaired,tan2017artgan} manually select subsets of those datasets for evaluating artwork synthesis.
But the curated datasets still suffer from several limitations: 
(a) the long tail distribution over classes makes it hard to fairly evaluate the generation performance on each class. Moreover, the author distribution of the artworks are also long-tailed, reducing the diversity of the dataset, i.e., most artworks are from the several top authors. (b) The labels are noisy. The quality of images are not consistent and there are semantic near-duplicates in some datasets. (c) The data collection, annotation, and preprocessing are non-standardized and thus the qualities of some images and annotations may not be satisfactory.
\fig{wikiart_curated} illustrates the class-imbalanced problem and duplicate images from the scraped WikiArt dataset as an example.

\footnotetext{Duplicate images URLs: \url{https://www.wikiart.org/en/leonardo-alenza/the-viaticum-1840} and \url{https://www.wikiart.org/en/leonardo-alenza/per-diem}}

\subsection{Dataset Creation} \label{sec:dataset_curation}

\begin{figure}[t]
\centering
\includegraphics[width=\linewidth]{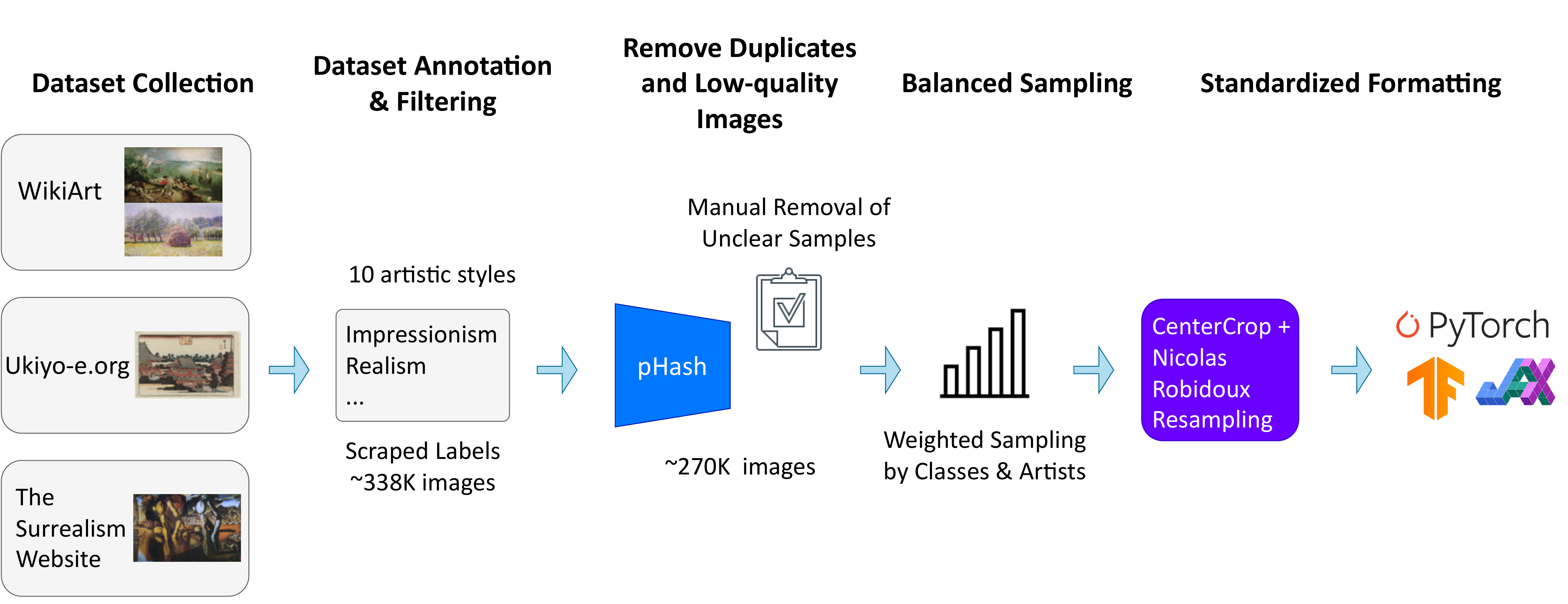}
\caption{ArtBench data collection process.}
\label{collection}
\end{figure}

To overcome the problems with previous artwork datasets, we design a standardized pipeline for data collection, annotation, filtering, sampling, and preprocessing, as shown in \fig{collection}. The standardized pipeline leads to our class-balanced, high-quality, cleanly-labeled, and standardized ArtBench-10 dataset.

\paragraph{Data collection} %
We collect raw images from three online databases with Fair Use licenses, namely Ukiyo-e (\url{https://Ukiyo-e.org}), WikiArt (\url{https://WikiArt.org}), and Surrealisum (\url{https://Surrealism.website}).
We then select artworks from the Metropolitan Museum of Art (MET), Museum of Fine Arts (MFA), Minneapolis Institute of Arts (MIA) and Fine Arts Museums of San Francisco (FAMSF) from the Ukiyo-e database, and all artworks from the other two databases.

\paragraph{Data annotation and filtering} We select the 10 artistic style categories with the highest frequencies in WikiArts. Then we
annotate each image with one artistic style. The Ukiyo-e and Surrealism are single-style databases. So we label all images from Ukiyo-e database as ``Ukiyo-e'' artistic style, and label all images from Surrealism database as ``Surrealism'' artistic style.
For the artworks from WikiArt, we extract the style information from the Alt-text HTML Attributes, and filter out the artworks that do not match any of the 10 pre-defined style categories. If an image matches with more than one artistic style, we heuristically select the first one that appears in the text description in the raw HTML data.
After the data annotation and filtering, we obtain a dataset containing 338K images.

\paragraph{Filtering out near-duplicates and low-quality images} 
Existing scrapings of WikiArt (see supplementary materials) contain near-duplicate artworks (replica painting, photography of exhibits, different digitization at different places, etc.), which harms the reliability of the benchmark. We filter out the near-duplicates with the perceptual hashing approach \cite{krawetz_2011}, resulting in 270K non-duplicate images.
We further filter out the low-quality images with low resolution or extreme aspect ratios, and the filtering criteria are described in the supplementary materials.

\paragraph{Balanced Sampling}
In order to create a class-balanced dataset, we aim to sample 6000 images per artistic style.
We also desire a diverse dataset, so we design a weighted sampling scheme, where the sampling weight is defined as the inverse frequency of the artist of the artwork. As a result, an artwork whose artist has more artworks is less likely to be sampled, and an artwork whose artist has fewer artworks is more likely to be sampled. With our weighted sampling scheme, we create a class-balanced and artist-balanced artwork dataset with maximum diversity.
We finally manually inspect the dataset and replace the low-quality samples (imperfect lighting, non-artworks, etc.) with the high-quality ones.
The dataset is randomly splitted into 50,000 training images and 10,000 testing images, with 5,000 training and 1,000 testing for each artistic style category.

\paragraph{Standardized formatting.}
To facilitate benchmarking different artworks synthesis approaches, we provide three versions of the dataset with different image resolutions: 32$\times$32, 256$\times$256, and original image sizes.
For the 32$\times$32, 256$\times$256 variants, we apply center-crop on the origianl images and then apply the Nicolas Robidoux resampling method \cite{robidoux_2012} to reduce artifacts during downsampling.
The three versions of our ArtBench-10 dataset are in the format of CIFAR (32$\times$32, tar archive), ImageFolder (256$\times$256, folders), and LSUN (origianl image sizes, LMDB file), so that they can be easily incorporated for different machine learning frameworks and image synthesis codebases.

\subsection{Dataset Statistics and Analysis}

\begin{figure}[t]
\begin{subfigure}{.65\linewidth}
  \centering
    \includegraphics[width=.9\linewidth]{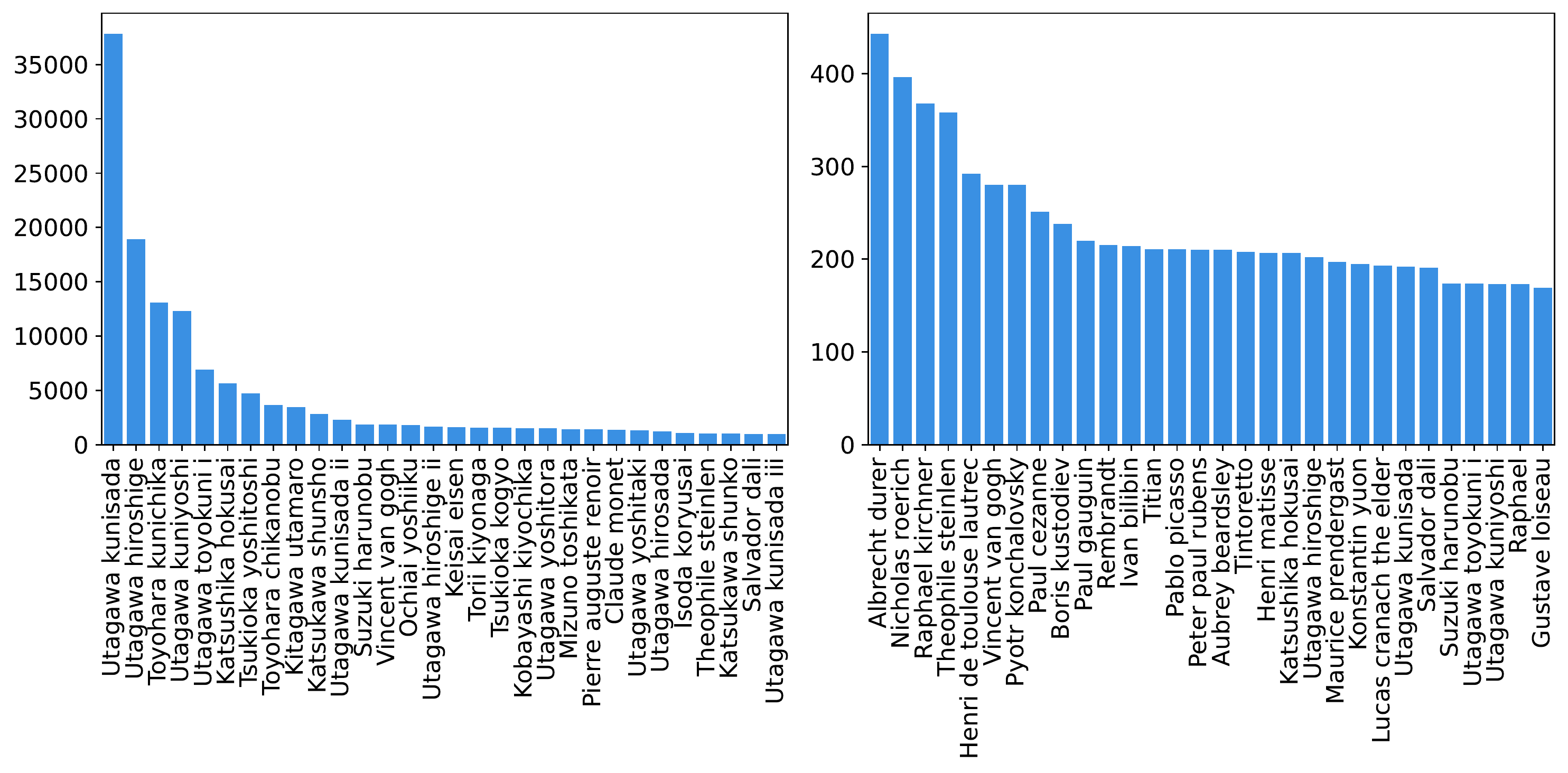}
    \caption{Number of artworks from artists in raw data (left) and ArtBench-10 (right). ArtBench-10 is more balanced on the distribution of artists.}
    \label{fig:sub2}
\end{subfigure}
\begin{subfigure}{.30\linewidth}
  \centering
    \includegraphics[width=.93\linewidth]{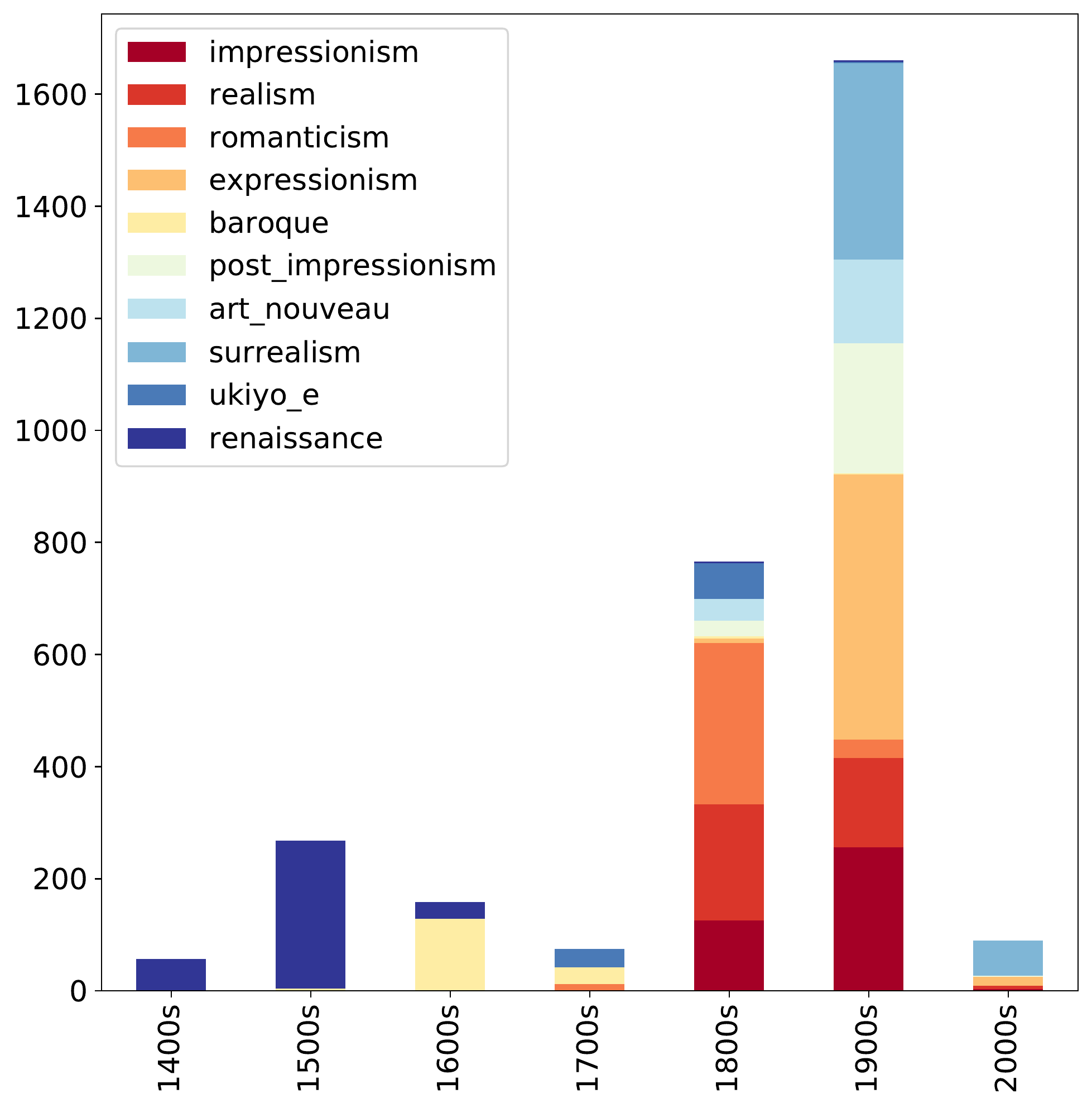}
    \caption{The distribution of the time periods of artists and styles. }
    \label{fig:sub3}
\end{subfigure}
    \caption{Statistics for ArtBench-10. Our dataset covers a wide of artworks across different time periods and greatly reduces imbalanced among artists for diversity.}
    \label{stats}
\end{figure}

\begin{figure}[t]
\begin{subfigure}{.48\linewidth}
  \centering
    \includegraphics[width=.9\linewidth]{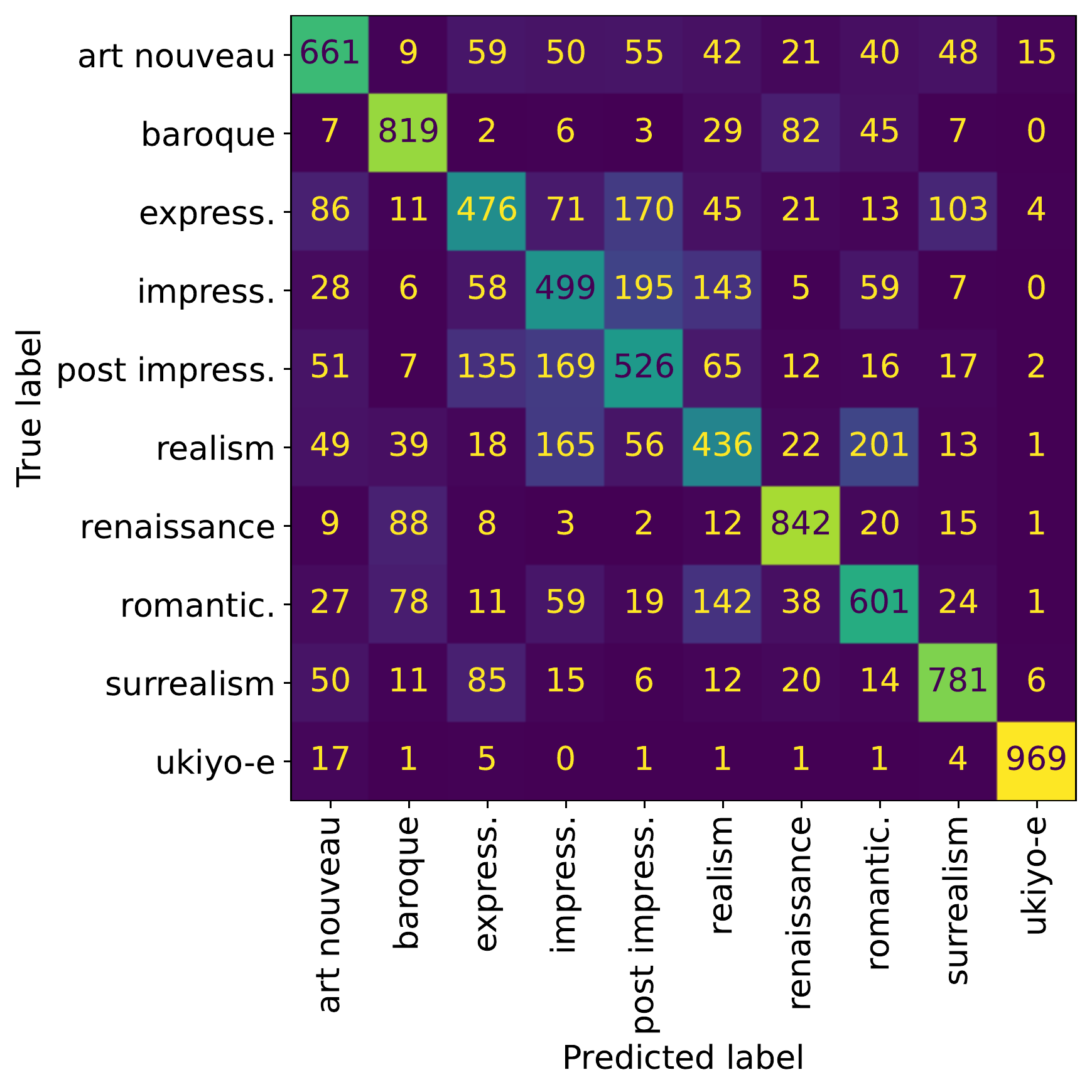}
  \label{fig:sub1}
\end{subfigure}%
\begin{subfigure}{.5\linewidth}
  \centering
    \includegraphics[width=.85\linewidth]{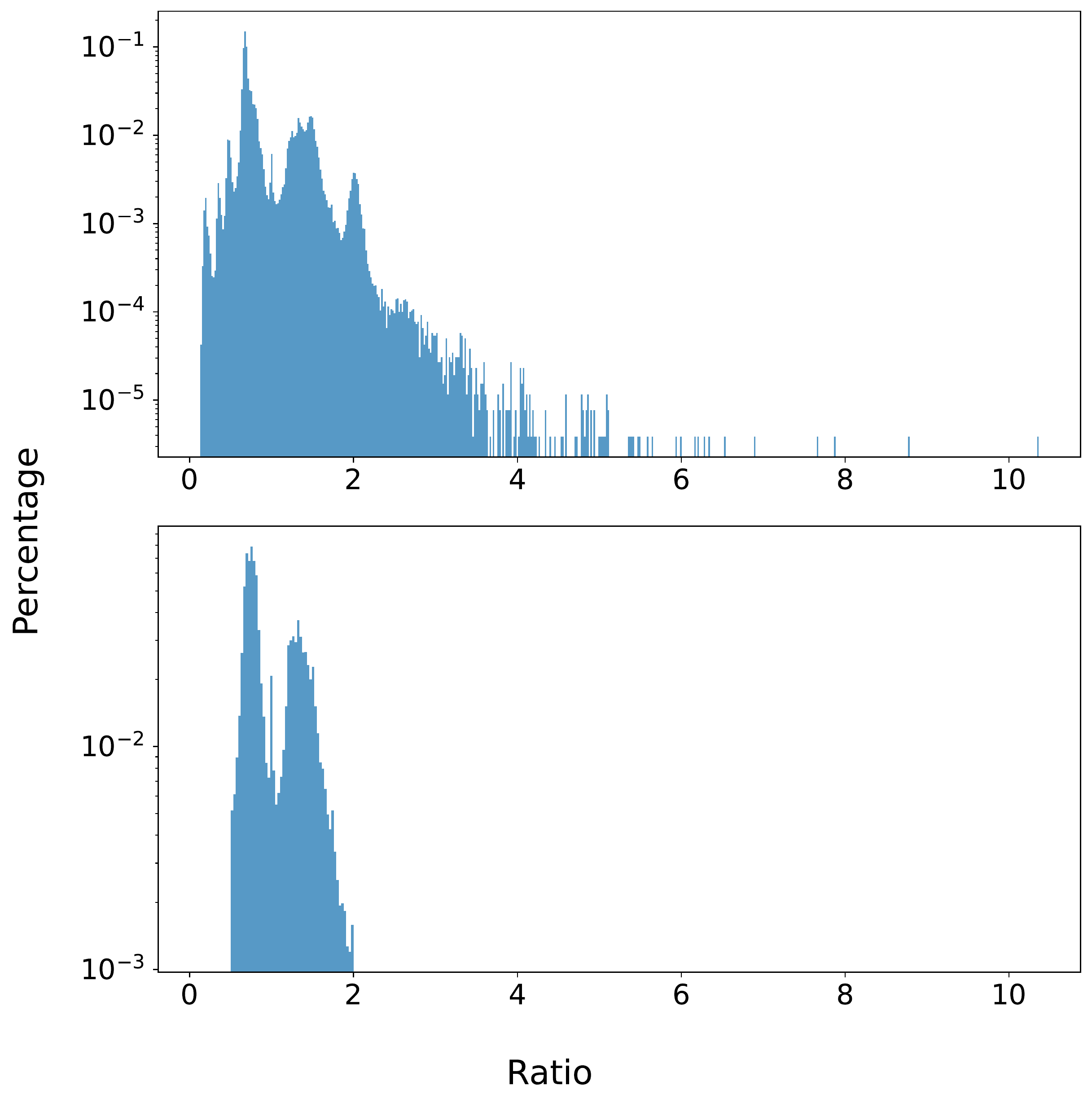}
  \label{fig:sub2}
\end{subfigure}
\vspace{-5pt}
\caption{(Left) Confusion matrix on the test set with SEResNet-34 trained on ArtBench-10 ($256 \times 256$);  (Right) Distribution of aspect ratios from raw data (top) and ArtBench-10 (bottom), respectively.}
\vspace{-10pt}
\label{cls}
\end{figure}

ArtBench-10 is a dataset of 60,000 artworks from 10 distinctive artistic styles. There are 5,000 images for training and 1,000 images for testing for each artistic style. The balanced class distribution makes this dataset suitable for benchmarking image synthesis.
Our balanced sampling scheme also guarantees that the artists of the artworks are balanced. \fig{stats}(a) demonstrates the author distribution before and after the balanced sampling.
The artworks in ArtBench-10 are from European, modern North American and East Asian ranging from 14th to 21st century.
\fig{stats}(b) illustrates the distribution of the time periods of artists and styles. 
The distribution of aspect ratios of the raw images and ArtBench-10 images is demonstrated in \fig{cls}(right).
It is clear that the ArtBench-10 dataset is filtered with with aspect ration thresholds [0.5, 2], while the raw dataset before filtering contains images with extreme aspect rations which are not suitable for training generative models.

We train an image classification model with SEResNet-34 on the $256\times256$ images of ArtBench-10.
\fig{cls}(left) shows the confusion matrix on the test set, indicating clear inter-class separation. %

\vspace{-10pt}
\section{Experiments}\label{sec:exp}
\vspace{-5pt}

In this section, we describe the experiment setup, evaluation metrics, quantitative and qualitative results, as well as further analysis.

\vspace{-5pt}
\subsection{Experiment Setup}\label{subsec:setup}
\vspace{-5pt}
We benchmark previous methods with four experimental setups: class-conditional synthesis with resolutions of $32\times32$, unconditional synthesis with resolutions of $32\times32$, class-conditional synthesis with resolutions of $256\times256$, and unconditional synthesis on the ``\textit{Impressionism}'' class with resolutions $256\times256$. The implementation details are provided in the supplementary materials.

We select representative image synthesis methods for benchmarking, including GAN-based methods~\cite{karras2020training, kang2021rebooting, Sauer2021NEURIPS, brock2018large}, VAE~\cite{vahdat2020NVAE}, and diffusion-based models~\cite{song2020denoising, nichol2021improved, liu2022pseudo}.

\vspace{-5pt}
\subsection{Evaluation Metrics} \label{sec:eval}
\vspace{-5pt}
For evaluation, we randomly sample 50,000 images for the unconditional synthesis setup, and sample 5,000 images for each class (leads to 50,000 images in total) for the conditional synthesis setup.
We adopt the commonly used evaluation metrics Inception Score (IS)~\cite{salimans2016is}, Fréchet Inception Distance (FID)~\cite{heusel2017fid}, Improved Precision and Recall~\cite{kynkaanniemi2019improved}, and Kernel Inception Distance (KID)~\cite{binkowski2018demystifying} for evaluation.

\vspace{-8pt}
\paragraph{Inception Score}
Inception Score (IS) \cite{salimans2016is} measures the ability of a generative model to capture the whole data distribution and producing high quality samples for each single class. 
However, IS does not take the whole data distribution into consideration and does not reflect the diversity of generated images.

\vspace{-8pt}
\paragraph{Fréchet Inception Distance}
Fréchet Inception Distance (FID) \cite{heusel2017fid} measures the distance between ground-truth and generated image feature distributions.
Compared to IS, FID leverages the dataset information and is considered to be more consistent with the noise level and human perception.

\vspace{-8pt}
\paragraph{Improved Precision and Recall}
\citet{kynkaanniemi2019improved} claims that IS and FID ignore the trade-off between the sample quality and diversity of the generated samples, and proposes improved precision and recall to overcome this issue. Precision is the percentage of generated images that fall into the estimated manifold of real images . Recall is the percentage of real images that fall into the estimated manifold of generated images. Precision measures the quality of generated images and recall measures the diversity.

\vspace{-8pt}
\paragraph{Kernel Inception Distance}
Kernel Inception Distance (KID) \cite{binkowski2018demystifying} measures the maximum mean discrepancy (MMD) on images with a kernel function. %
KID shares some nice properties of FID such as being capable of reflecting the artifacts of images. It can also compare skewness between distributions and is an unbiased estimator.

\vspace{-5pt}
\subsection{Quantitative Results} 
\vspace{-5pt}

We conduct extensive experiments with both conditional and unconditional setups and different image resolutions following the experiment setups described in~\sect{subsec:setup}. The quantitative results as shown in~\tbl{perf} and~\tbl{perf_256}.
StyleGAN2 + ADA outperforms all other models under both conditional and unconditional setups in terms of all metrics except for Precision with images of different resolutions.
In~\tbl{perf}, note that the precision scores of StyleGAN2 + ADA are relatively low, while Projected GAN and Improved DDPM have the highest precision.
We note that the performance gap between StyleGAN2 + ADA and Projected GAN is much more prominent under higher resolutions.

\begin{figure}[H]
\includegraphics[width=\linewidth]{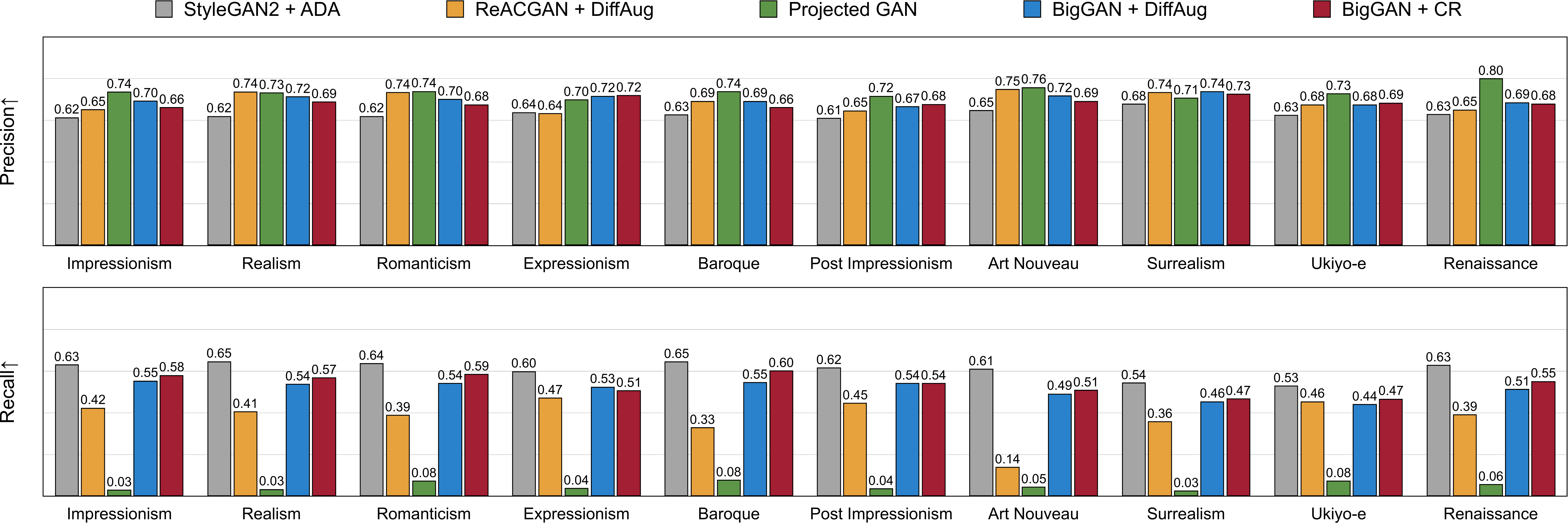}
\caption{Per-class precision and recall for conditional models trained on ArtBench-10 ($32 \times 32$).}
\label{per_class_stat}
\end{figure}

We also show the per-class precision and recall in \fig{per_class_stat} to demonstrate which classes are easier and which are harder.
It is shown that Projected GAN achieves consistently low recall's across different classes, indicating that the images generated by Projected GAN are of low diversity.

\begin{table}[h]
  \centering
  \resizebox{\textwidth}{!}{
  \begin{tabular}{lllllll}
    \toprule
    Setting & Model     &  FID$\downarrow$     & IS$\uparrow$ ($\pm \sigma$) & Precision$\uparrow$ & Recall$\uparrow$ & KID$\downarrow$ \\
    \midrule
    $32\times32$, Conditional
    & StyleGAN2 + ADA \cite{karras2020training} & $\mathbf{2.625}$  & $\mathbf{6.502} \pm 0.053$ & $0.652$ & $\mathbf{0.593}$ & $\mathbf{0.00066}$ \\
    & StyleGAN2  \cite{karras2020analyzing} & $4.491$  & $6.262 \pm 0.0933$ & $0.671$ & $0.534$ & $0.00155$     \\
    & ReACGAN + DiffAug \cite{kang2021rebooting, zhao2020differentiable}     &  $3.175$ & $5.240 \pm 0.522$ & $0.694$ & $0.513$ & $0.00098$  \\
    & Projected GAN \cite{Sauer2021NEURIPS}     & $11.837$ & $4.991 \pm 0.442$ & $\mathbf{0.727}$ & $0.119$ & $0.00275$     \\
    & BigGAN + DiffAug \cite{brock2018large, zhao2020differentiable} & $4.055$  & $5.827 \pm 0.559$ & $0.707$ & $0.508$ & $0.00166$     \\
    & BigGAN + CR \cite{brock2018large, Zhang2020Consistency}  & $4.647$  & $5.745 \pm 0.501$ & $0.695$ & $0.538$ & $0.00192$   \\
    \midrule
    $32\times32$, Unconditional & StyleGAN2 + ADA \cite{karras2020training} & $\mathbf{2.783}$  & $\mathbf{6.510} \pm 0.0736$ & $0.656$ & $\mathbf{0.580}$ & $\mathbf{0.000754}$    \\
    & StyleGAN2 \cite{karras2020analyzing} & $3.936$  & $6.336 \pm 0.0615$ & $0.660$ & $0.551$ & $0.00116$    \\
    & S-PNDM \cite{song2020denoising, liu2022pseudo} & $16.677$ & $6.390 \pm 0.0505$ & $0.666$ & $0.546$ & $0.0108$ \\
    & F-PNDM \cite{song2020denoising, liu2022pseudo} & $16.196$ & $6.389 \pm 0.0923$ & $0.665$ & $0.543$ & $0.01051$ \\
    & FON \cite{song2020denoising, liu2022pseudo} & $16.770$ & $6.440 \pm 0.0478$ & $0.670$ & $0.540$ & $0.0111$  \\
    & DDIM \cite{song2020denoising, liu2022pseudo} & $17.561$ & $6.456 \pm 0.123$ & $0.673$ & $0.544$ & $0.0118$  \\
    & PF \cite{song2020denoising, liu2022pseudo} & $19.653$ & $6.487 \pm 0.0770$ & $0.665$ & $0.530$ & $0.0134$ \\
    & Improved DDPM \cite{song2020denoising, nichol2021improved} & $15.308$ & $5.496 \pm 0.0356$ & $\mathbf{0.763}$ & $0.396$ & $0.0119$ \\
    & NVAE \cite{vahdat2020NVAE} & $88.248$ & $4.438 \pm  0.0415$ & $0.625$ & $0.0988$ & $0.0812$  \\
    \bottomrule
  \end{tabular}
  }
  \vspace{0.5em}
   \caption{Benchmarking different image synthesis methods on $32\times32$ ArtBench-10 with both class-conditional and unconditional setups.}
   \vspace{-5pt}
   \label{perf}
\end{table}

\begin{table}[H]
  \centering
  \resizebox{\textwidth}{!}{
  \begin{tabular}{lllllll}
    \toprule
    Setting & Model     &  FID$\downarrow$     & IS$\uparrow$ ($\pm \sigma$) & Precision$\uparrow$ & Recall$\uparrow$ & KID$\downarrow$ \\
    \midrule
    $256\times256$, Conditional & Projected GAN \cite{Sauer2021NEURIPS} & $42.474$ & $5.423 \pm 0.0743$ & $\mathbf{0.621}$ & $0.0356$ & $0.00595$ \\
    & StyleGAN2 + ADA \cite{karras2020training} & $\mathbf{7.550}$  & $6.634 \pm 0.149$ & $0.606$ & ${0.251}$ & $\mathbf{0.00190}$ \\
    & Latent Diffusion \cite{rombach2021highresolution} & $17.577$ & $\mathbf{7.780} \pm 0.113$ & $0.576$ & $\mathbf{0.356}$ & $0.00655$ \\
    \midrule
    $256\times256$, Impressionism, Unconditional & Projected GAN \cite{Sauer2021NEURIPS} & $\mathbf{11.309}$ & $\mathbf{8.026} \pm 0.120$ & $0.612$ & $0.360$ & $\mathbf{0.00190}$ \\
    & Latent Diffusion \cite{rombach2021highresolution} & $20.535$ & $6.463 \pm 0.107$ & $\mathbf{0.652}$ & $\mathbf{0.363}$ & $0.00991$ \\
    \bottomrule
  \end{tabular}
  }
  \vspace{0.5em}
   \caption{Benchmarking different image synthesis methods on $256\times256$ ArtBench-10 with both class-conditional and unconditional setups.}
   \vspace{-5pt}
   \label{perf_256}
\end{table}

\vspace{-10pt}
\subsection{Qualitative Results}
\vspace{-5pt}
\begin{figure}[t]
    \centering
    \includegraphics[width=\linewidth]{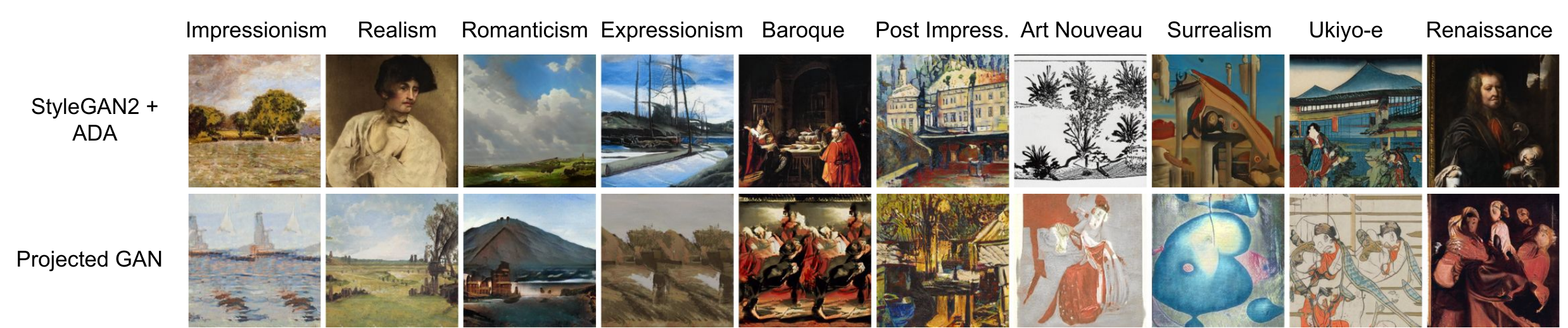}
    \caption{Generated samples per style from StyleGAN2 + ADA~\cite{karras2020training} and Projected GAN~\cite{Sauer2021NEURIPS} trained on ArtBench-10 ($256\times256$).
    }
    \vspace{-5pt}
    \label{band}
\end{figure}

\begin{figure}[H]
\centering
\includegraphics[width=\linewidth]{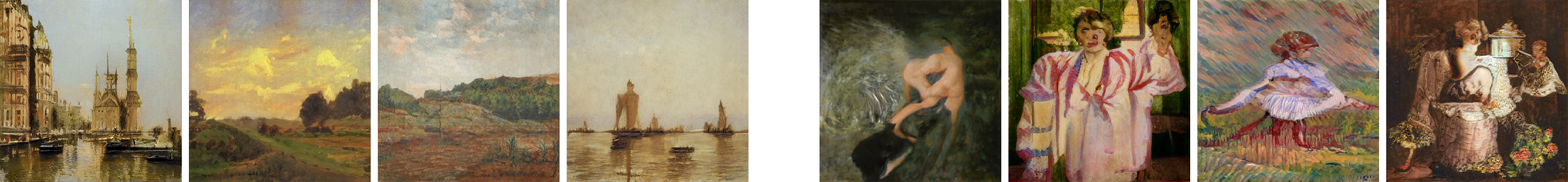}

\caption{Success (left) and failure (right) cases of Projected GAN~\cite{Sauer2021NEURIPS} on ArtBench-10-Impressionism ($256\times256$). Projected GAN is good at landscapes art, but struggles for portraits.}
\label{qual}
\vspace{-20pt}
\end{figure}

\fig{band} shows generated samples for each artistic style by StyleGAN2 + ADA and Projected GAN.
\fig{qual} demonstrates several success and failure cases by a representative generative model, Projected GAN \cite{Sauer2021NEURIPS}.
The current generative models are good at generating paintings of landscape, cityscape, and marina artworks, but struggles to generate portraits. This observation is consistent with many other works on generative models.
More qualitative results of other generative models are demonstrated in the supplementary materials.

\begin{figure}[H]
\centering
\includegraphics[width=\linewidth]{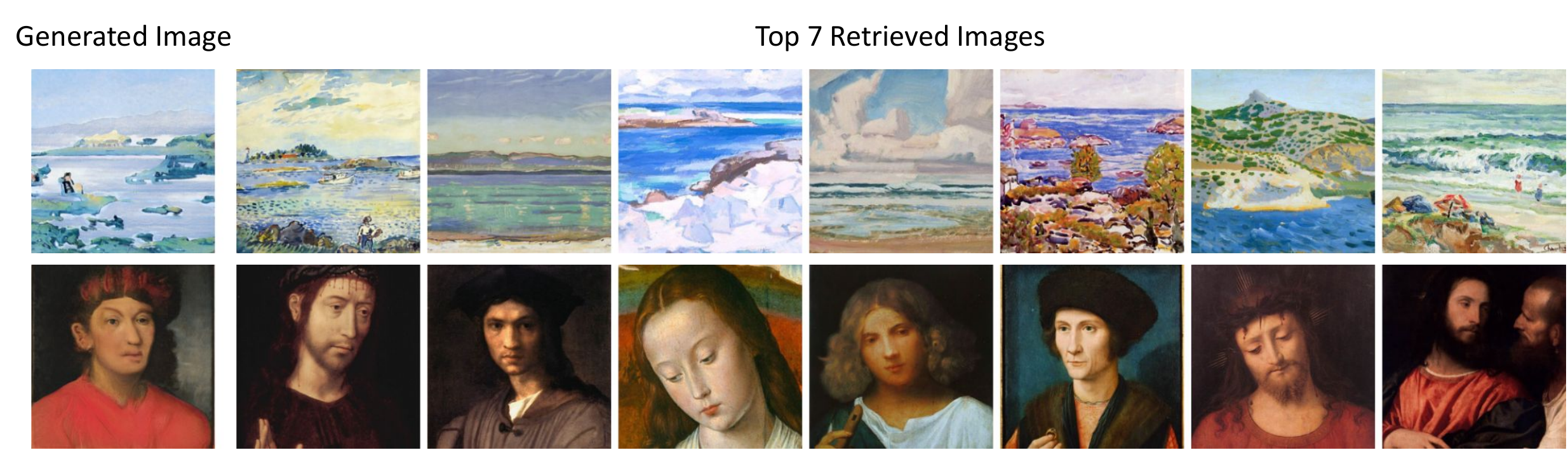}
\caption{Top-7 retrieval results for StyleGAN2 + ADA \cite{karras2020training} ($256\times256$)
. The first column is generated images as queries, and the other columns are retrieved images from training set.}
\vspace{-5pt}
\label{knn}
\end{figure}

We further conduct a nearest neighbor retrieval experiment with the $256\times256$ class-conditional StyleGAN2 + ADA model. Given a generated image, we retrieve its 7 nearest neighbors from the training set. The features for retrieval are extracted by an ImageNet-pretrained ResNet-50 model. Results in~\fig{knn} demonstrate that the generative model is not simply memorizing images from the training set, but rather has some extent of extrapolation and is able to create new samples that it has not seen in the training set.

\vspace{-5pt}
\section{Conclusion and Discussions}\label{sec:conclusion}
\vspace{-5pt}
In this paper, we propose ArtBench-10, the first class-balanced, high-quality, cleanly-annotated, and standardized benchmark for artworks generation.
With standardized data collection, annotation, filtering, and preprocessing processes, ArtBench-10 addresses common problems in previous artwork datasets, such as label imbalance, near-duplicates, noisy labels, and poor image quality. 
We benchmark representative image synthesis models on ArtBench-10, and conduct analysis on the results.
We hope that ArtBench-10 will set a standard benchmark for future work on artwork generation.

Due to the artwork databases we choose, ArtBench-10 is heavily biased towards artworks from European, modern North American and East Asian. %
We plan to address this issue in future works by incorporating artworks from larger geographical ranges and longer time periods, especially from Asia, Africa, South America and Oceania.
We also acknowledge the potential risk that the dataset could be exploited to create forgery artworks and cause copyright violations beyond Fair Use. 

\begin{ack}
The authors would like to thank Keyulu Xu, Ruijie Fang and Wendi Yan for their support
and help, Zeyu Cao for some of the computational resources, as well as an art history researcher (who would like to remain anonymous) for valuable discussions during the earlier stages of this work. The authors would also like to express gratitude towards WikiArt, the Surrealism Website, Japanese Woodblock Print Search, and all the artists who appeared in the dataset. Peiyuan Liao is also affiliated with Praxis Pioneering for works in machine learning.
\end{ack}

\newpage

\bibliography{refs}
\bibliographystyle{plainnat}

\appendix
\newpage

\section{Appendix}

\subsection{Datasheet for ArtBench-10}

We present the documentation and intended uses of ArtBench-10, as well as means to access it, in the form of a datasheet following \citet{51056} and \citet{gebru2021datasheets}. The dataset is available through a GitHub repository (\url{https://github.com/liaopeiyuan/artbench}) as well as a long-term Kaggle dataset (\url{ https://doi.org/10.34740/KAGGLE/DS/2250482}). %

\subsubsection{Motivation} \label{motivation}

\paragraph{Who funded the curation of the ML dataset, and for what purpose? Who was involved in the curation process? Please list all individuals and/or institutions.} \

The dataset is authored by 4 researchers distributed across 2 institutions: Peiyuan Liao (Carnegie Mellon University), Xiuyu Li (UC Berkeley), Xihui Liu (UC Berkeley) and Kurt Keutzer (UC Berkeley).

The authors describe the dataset as the first class-balanced, high-quality, cleanly annotated and standardized dataset for benchmarking artwork generation.

\paragraph{If the ML dataset was sourced from an underlying curated source (e.g. a museum), please outline the factors governing the inclusion of artworks into the source collection.} \

Images are sourced from WikiArt, the Surrealism Website and Japanese Woodblock Print Search
(Ukiyo-e). WikiArt is a community project, where various contributors across the globe can add to the collection. On the other hand, Ukiyo-e is created by John Resig (partly sponsored by Ritsumeikan University) and Surrealism is created by Adam McLean, implying that the works are selected by these individuals.

\paragraph{If the ML dataset was sourced from an underlying curated source, please outline the factors motivating the choice of the data source.} \

The datasets are chosen mainly due to the abundance of images belonging to certain art styles, since more artworks leaves space for more effective filtering techniques that generates a higher quality dataset designed for machine learning use cases.

\subsubsection{Data Provenance} \

\paragraph{Is there a provenance policy outlining the ethics and conditions of transfer of works in the collection? For example, were the visual artworks or music pieces obtained by means of transfer
from other sources? Please describe the terms and conditions of provenance policy.} \

There isn't a formal provenance policy outlining the ethics and conditions of transfer of works in the collection.
The artworks from Wikiart are governed by the following terms of use: \url{https://www.wikiart.org/en/terms-of-use}; for the Surrealism website, no terms and conditions are supplied; for Ukiyo-e, the terms and conditions are outlined in the respective sources.

\paragraph{Does the dataset include ownership histories, condition of the artworks at the time of procurement/inclusion into the dataset, and legal documentation related to the transfers? Please
describe any available information related to ownership histories.} \

ArtBench-10 does not include such inoformation. However, minimal details relating to histories of artworks are captured in the respective data sources that comprise ArtBench-10.

\paragraph{What type of information about the music piece/visual artwork’s attribution and origination
is included? Is it limited to a single creator, or does the dataset afford other categories such
as the name of the person who commissioned the artwork, workshop assistants, symphony
performers’ names, its owners or stewards over time, names of remixers or transliterators,
etc.? Please list any available information concerning the people associated with the artwork.} \

The information pertaining to artwork’s attribution is limited to the artist.

\paragraph{From which source(s) does the dataset draw authenticity verification and title of artworks and
music pieces? Please name specific cultural institutions, private or public repositories, and all
other sources. Please describe the types of personnel, organizations, and processes involved in
the production of authentication and title information.} \

Authenticity verification of the Surrealism Website and Ukiyo-e are done by the respective individuals (see \ref{motivation}). For WikiArt, artworks are collected from museums, universities, town halls, and other civic buildings of more
than 100 countries by users (\url{https://www.wikiart.org/en/How-to-Contribute}), and each uploader bears responsibility for the authenticity (\url{https://www.wikiart.org/en/terms-of-use}).

\paragraph{Is information pertaining to the music piece/visual artwork’s appreciation included in the
dataset? For example, information about multiple interpretations of an artwork/music piece,
reviews, labels imprinted on the artworks/music pieces, etc. Please describe.} \

All of Wikiart, Surrealism and Ukiyo-e do not include original interpretations of the artworks.

\paragraph{Are the values of Indigenous and minoritized groups honored while including the visual
artwork/music piece in the dataset? Describe the process or protocol by which visual artworks/music pieces affiliated with these groups have been approved for use in the dataset.} \

Indigenous artifacts and artifacts from minoritized groups are underrepresented in ArtBench-10, as it is heavily biased towards artworks from European, modern North American and East Asian cultures. ArtBench-10 contains both public domain and
copyright protected artworks, the latter of which are released under a Fair Use license (\url{https://www.wikiart.org/en/about} for details). Each image in ArtBench-10 is correspondingly tagged with whether it belongs to public domain in the metadata \cite{artbench_kaggle}.

\subsubsection{Data Composition}

\paragraph{What real-world objects or phenomena do the data instances represent (e.g., material objects,
performances)?} \

The data instances are images of artworks, such as paintings, sculptures, and
photographs, across a limited set of media.

\paragraph{If the dataset has been derived from a larger curated corpus, please describe the relationship
between the dataset and the larger sample(s) by explaining environmental, sociopolitical, and
professional conditions that presently or historically affect the gap between the dataset and
total corpus.} \

See Section \ref{sec:dset} in the main text for details of sample selection.

\paragraph{How many instances are there in total, including number of instances per type (e.g., number
of artworks/music pieces per artist )? Are there recommended data splits (e.g., training, development/validation, testing) based on the number of instances per type? Please provide a
description of these splits, explaining the rationale behind them.} \

See Section \ref{sec:dset} in the main text for details of data splits and distribution of artworks.

\paragraph{Is the dataset limited by intellectual property restrictions imposed by an external resource (e.g.
a cultural institution’s decision to exclude works not in the public domain)? Please provide
descriptions of external resources and any associated restrictions, including links to access
points and policies as relevant.} \

ArtBench-10 contains both public domain and
copyright protected artworks, the latter of which are released under a Fair Use license (\url{https://www.wikiart.org/en/about} for details). Each image in ArtBench-10 is correspondingly tagged with whether it belongs to public domain in the metadata \cite{artbench_kaggle}.

\paragraph{If data instances are proxies for visual objects, is there information about how the objects
were converted to images? E.g., type of camera/model, camera parameters, lighting conditions,
etc.? And is there a justification for the choice of the specific viewpoint, camera, and lighting
conditions? What aspects of the artwork/object may be missing as a consequence of the specific
viewpoint/lighting conditions? Please describe.} \

No details about the method of digitization are provided.

\paragraph{If data instances are proxies for musical pieces, is information available about how performances were recorded? E.g. type of equipment, studio or live audience setting? Which aspects
of the performance may be missing as a consequence of recording conditions? Please describe.} \

N/A

\paragraph{If data instances are proxies for visual objects or fashion pieces, which view(s) of the original
objects do data instances represent? E.g., top view, side view, etc.?} \

Data instances represent the frontal view, and images are available at multiple resolutions (32x32, 256x256, and original size).

\paragraph{Does the dataset contain opinions, sentiments, or beliefs that relate to specific social or cultural groups (e.g., religious beliefs, political opinions, etc.) or visualizations/musical renderings
that might elicit feelings of anxiety or distress (e.g., those depicting wars, oppression, etc.) ? If
so, please describe.} \

ArtBench-10 may contain sensitive information like those concerning religious beliefs
and political opinions, and artworks that might elicit feelings of anxiety or distress.

\subsubsection{Data Collection}

\paragraph{How were the individual visual artworks or music pieces collected? Were they obtained directly from museums, galleries, music labels, archives, etc.? Or were they derived from existing ML datasets? Please provide details concerning the data acquisition process} \

60,000 individual visual artworks are collected from the three databases, namely WikiArt, Ukiyo-e and Surrealism. They were scraped form the official websites, which originates from museums, universities,
town halls, private collections, and other civic buildings of more than 100 countries.

\paragraph{Is there any institutional (e.g., museum, music label, etc.) policy governing the data collection process? Please describe any applicable institutional ethical standards/rules followed in
collecting the data.} \

Such standards are not available.

\paragraph{Are all relevant stakeholders (e.g. artists, auction houses, galleries, museums, music labels,
composers, legal experts, archivists, etc.) consulted in the collection process? For example,
how has the dataset collection process been informed by art conservationists, art historians,
and other stakeholders (e.g., Indigenous groups) in order to ensure cultural relevance and
longevity? Please describe.} \

Only the paper authors are involved in the data collection process.

\paragraph{Is there information on the IP rights and cost of acquisition of an artwork or a music piece? 
Is there information pertaining to their usage, display, reuse, sale, reproduction, and transformation? Please describe all relevant terms and conditions.} \

There is no IP rights and cost of acquisition of individual artworks beyond the ones already committed by the maintainers of the respective databases ArtBench-10 is built from.

\paragraph{Is there permission to collect/view/reproduce culturally sensitive information? For e.g., those
pertaining to Indigenous and minoritized groups, secret objects, sacred objects, etc. that may
be forbidden from public view? Please describe the details related to consent request and
approval from the concerned individuals and/or communities.} \

Such permissions are governed by the rules of the respective data sources, under a Fair Use license.

\paragraph{Do the included works provide wide ranging and multi-dimensional narratives? Please describe how diversity and inclusion has been ensured in the data collection process—for example, demonstrate that the collection is not skewed by providing the number of works across
different artists, sources, communities, and geographies; describe how the collection upholds
ethical values of the groups it includes and showcases diverse perspectives.} \

See Section \ref{sec:dset} in the main text for details on distribution of artists and time periods.

\subsubsection{Data Pre-processing, cleaning, and labeling}

\paragraph{Were there any specific pre-processing steps (e.g. filtering, resizing, rotations, conversion from
color to grayscale, etc.) that might have resulted in loss of information? Please describe} \

See Section \ref{sec:dset} in the main text for details on preprocessing steps.

\paragraph{Were there any translations involved in obtaining the relevant metadata? e.g. translating
description of an aboriginal artwork from a native language to English? If so, please describe
the diligence process to ensure accurate translations.} \

No translations are involved.

\paragraph{Does the dataset identify any subpopulation (e.g. by ethnicity, gender expression) by virtue of
text inscribed on the artwork or its frame? If so, describe whether such terms are captured
in the annotation process and the method of transcription. Describe how such terms are interpreted in their original historical context and how the dataset’s users and reviewers can trace
the terms.} \

The dataset does not identify any subpopulation by virtue of text inscribed on the artwork or its frame.

\paragraph{Does the dataset identify any subpopulation in languages other than modern English (e.g. ancient Greek, modern French, modern creole and pidgin languages) or scripts other than standard English characters (e.g., types of Arabic script)? Describe how the dataset documents
subpopulations by language(s) and script(s).} \

The dataset does not identify subpopulation in languages other than modern English.

\paragraph{Is there a policy/guideline related to the way labeling/annotation was done? For e.g., any
(minimum and maximum) limit on the number of words/characters to describe the artwork,
the relevant contexts (cultural, historical, social, etc.) the annotation should cover. Any specific
protocol to label based on the type of artwork? Please describe. If the labels were already
obtained from existing GLAMs, is there documentation pertaining to how the labels were
created? For example, were there any institution specific guidelines? Please describe.} \

The labeling of Surrealism and Ukiyo-e are done by their respetive curators, and the labeling of WikiArt is done by their contributors, outlined in the contribution guideline (\url{https://www.wikiart.org/en/How-to-Contribute}).

\paragraph{What information is included in the labels? For example, artist’s name, artist’s nationality,
date associated with the artwork, associated art movement (Byzantine, Renaissance, etc.),
genre (classical music, portraits, etc.), art material (oil, paint, etc.), source of the artwork
(museum name, name of the recording label, etc.), etc. Please describe.} \

Art style is the only information included in the labels. However, the metadata contains additional information such as artist name, length and width, whether the image is in public domain, and original URL.

\paragraph{Is there sufficient provenance information (e.g. historical, political, religious, cultural evidence,
artist, date, art period, etc.) pertaining to the artwork to generate reliable labels? If not,
describe how the labels were obtained/created?} \

Some data provenance information are available to the curators to generate reliable labels.

\paragraph{Describe the background of the annotators (e.g., education, geographic location, their sociocultural contexts, etc.) Do they possess relevant qualifications/knowledge to label the instance?
If so, please justify. Are there some aspects of their specific backgrounds (e.g. political affiliations, cultural contexts, etc.) that could have biased the annotations? If so, please describe.} \

\begin{enumerate}
    \item Ukiyo-e: John Resig is a Visiting Researcher at Ritsumeikan University in Kyoto working on the study of Ukiyo-e.
    \item Surrealism: Adam McLean is a Scottish writer on alchemical texts and symbolism. 
    \item WikiArt: there are no background information concerning the annotators.
\end{enumerate}

\subsubsection{Data use and distribution}

\paragraph{What are the terms and conditions information pertaining to the use and distribution of the
dataset? For example, are there any licenses, required clearances from ministries of culture?
Please describe the terms and conditions related to such license/clearance.} \

The data is released under a Fair Use license, and the code is released under a MIT license: \url{https://github.com/liaopeiyuan/artbench/blob/main/LICENSE}.

\paragraph{Is there a link or repository that lists all papers/works that have used the dataset? What are
some potential applications and scenarios where the dataset could be used?} \

The main text presents several baseline methods in the literature that can use ArtBench-10 to train and evaluate conditional and unconditional image synthesis models. In addition, ArtBench-10 can be used to perform image recognition and classification tasks.

The GitHub repository (\url{https://github.com/liaopeiyuan/artbench}) will aim to keep track of all that have used the dataset, as well as a Kaggle dataset website \cite{artbench_kaggle}.

\paragraph{If the data concerns people (such as Indigenous groups), have they provided consent for the
use and distribution of the dataset in the context of the specific application? If the dataset is
going to be distributed, please describe the diligence process for ensuring that there are no
existing restrictions on sharing the artworks and metadata.} \

Such processes are outlined in the respective licenses of the data sources, such as WikiArt (\url{https://www.wikiart.org/en/about}), Surrealism (\url{https://surrealism.website/}) and Ukiyo-e (\url{https://ukiyo-e.org/about}).

\paragraph{Are there any aspects of human rights and dignity that are prone to be suppressed in a specific
use or distribution of the dataset? Please describe applications and scenarios where the dataset
should not be used or distributed.} \

Subjects portrayed in the artworks may not have provided consent during the creation of the work, and may be inappropriate for the general audience. The works themselves may exhibit negative social impacts, such as personally identifiable information, cultural appropriation, violence, slavery, or racism. It is also conceivable that the dataset could be exploited to create forgery, or other forms of copyright violation beyond Fair Use. We do not endorse these types of act, and we are neither responsible for the content nor the meaning of these images.

\paragraph{If the dataset will be distributed under a copyright or other intellectual property license,
and/or under applicable terms of use, please describe whose (individual or group’s) interest/stake in the data is likely to be promoted, and whose interest/stake is likely to be suppressed,
and why so. Please justify the use/distribution of the dataset given this context.} \

By releasing the code under a MIT license and the data under a Fair Use license, adopters of the dataset have freedom in interpreting the artworks and metadata, creating further distances between the creators and the users, and thus preventing potential promotion of certain author-affiliated interests. On the other hand, the choice of certain inclusion and exclusion of individual samples encourages groups that have interest to the message in a particular artwork while discouraging others.

Nevertheless, the authors and their institutions will benefit from the popularity of this dataset through the mechanism of academic citation in scholarly literature.

\subsubsection{Data Generation}

\paragraph{Is there any generated content in the dataset? If so, please explain the source data that is used
as examples to generate the desired output.} \ 

N/A

\subsubsection{Data Maintenance}

\paragraph{Who is maintaining the dataset and how can they be contacted?} \ 

The dataset will be maintained by Peiyuan Liao (Carnegie Mellon University), Xiuyu Li (UC Berkeley), Xihui Liu (UC Berkeley) and Kurt Keutzer (UC Berkeley). They can be contacted at \texttt{artbench.dataset@gmail.com}.

\paragraph{Is there a policy outlining the dataset update process? For example, will the dataset be updated
as and when there is a change in information with respect to a data instance or will the dataset
be updated at regular time intervals to reflect all changes in that time interval? Is there a
description of the specific changes made- e.g., provenance related, reason for removal of a
data instance, or addition of a data instance, etc. Please describe.} \ 

Currently, there are no concrete plans regarding the dataset update process, as the work is recently curated. The authors will ensure continued access to ArtBench-10 through hosting solutions including GitHub, Google Drive and Kaggle. The metadata and dataset content are not expected to change, and maintenance efforts are expected if unexpected changes did take place. 

\paragraph{Is there is a provision for others to contribute to the dataset? If yes, describe the process for
ensuring the authenticity of the information associated with the contributed data instances.} \ 

Currently, no other individuals may contribute to the dataset.

\subsection{Metadata Schema} \label{apppend:metadata}

A metadata table is also included in the release of ArtBench-10, presented as comma-separated values (CSV), and available at \citet{artbench_kaggle}. Certain fields are available for each sample, including (in JSON Table Schema):

\begin{lstlisting}[language=json,firstnumber=1]
{
"fields": [
{
	"name": "name",
	"title": "name",
	"constraints": {
		"required": true,
		"type": "http://www.w3.org/2001/XMLSchema#string"
	}
},
{
	"name": "artist",
	"title": "artist",
	"constraints": {
		"required": true,
		"type": "http://www.w3.org/2001/XMLSchema#string"
	}
},
{
	"name": "url",
	"title": "url",
	"constraints": {
		"required": true,
		"type": "http://www.w3.org/2001/XMLSchema#anyURI"
	}
},
{
	"name": "is_public_domain",
	"title": "is_public_domain",
	"constraints": {
		"required": true,
		"type": "http://www.w3.org/2001/XMLSchema#boolean"
	}
},
{
	"name": "length",
	"title": "length",
	"constraints": {
		"required": true,
		"type": "http://www.w3.org/2001/XMLSchema#int"
	}
},
{
	"name": "width",
	"title": "width",
	"constraints": {
		"required": true,
		"type": "http://www.w3.org/2001/XMLSchema#int"
	}
},
{
	"name": "label",
	"title": "label",
	"constraints": {
		"required": true,
		"type": "http://www.w3.org/2001/XMLSchema#string"
	}
},
{
	"name": "split",
	"title": "split",
	"constraints": {
		"required": true,
		"type": "http://www.w3.org/2001/XMLSchema#string"
	}
},
{
	"name": "cifar_index",
	"title": "cifar_index",
	"constraints": {
		"required": true,
		"type": "http://www.w3.org/2001/XMLSchema#int"
	}
}
]
}
\end{lstlisting}

\subsection{Weighted Sampling Algorithm}
This section describes how we sample 6000 images per artistic style from the filtered dataset, while having it to be artist-balanced. Let $\bm{S}$ be the set that contains all $10$ styles, and $\mathcal{M}_\alpha$ maps each style to the set of all images in the filtered dataset that belong to it, then images of style $s$ can be represented by $\mathcal{M}_\alpha(s)$, where $s \in \bm{S}$. Similarly, let $\mathcal{M}_\beta$ map each image to its corresponding artist, and let $\mathcal{M}^s_\gamma$ map each artist to the number of artworks in style $s$ created by them. The weighted sampling algorithm to get the final \name is shown in \algo{alg:sampling}, where the sampling weight is defined as the inverse frequency of the artist of the artwork per style.

\begin{algorithm}
\caption{Weighted Sampling} \label{alg:sampling}
\textbf{Input:} $10$ artistic styles $\bm{S}$, style to images map $\mathcal{M}_\alpha$, image to artist map $\mathcal{M}_\beta$\\
\textbf{Output:} the images in \name dataset $\bm{D}$
\begin{algorithmic}
\STATE $rand \gets$ {random sampling function that takes in $(n, \bm{X}, \bm{W})$ to randomly select $n$ samples without \hspace*{3.4em} replacement from $\bm{X}$ based on $\bm{W}$, which stores the weight for each $x \in \bm{X}$}
\STATE $norm \gets$ {function that normalizes a set by dividing each element by the sum of all elements} 
\STATE $\bm{D} \gets \emptyset$
\FOR{$s$ \textbf{in} $\bm{S}$}
    \STATE $\bm{X} \gets \mathcal{M}_\alpha(s)$
    \STATE $\bm{W} \gets \emptyset$
    \STATE $\mathcal{M}^s_\gamma \gets$ {artist to number of artworks map in style $s$}
    \FOR{$x$ \textbf{in} $\bm{X}$}
        \STATE $y \gets \mathcal{M}_\beta(x)$
        \STATE $\bm{W} \gets \bm{W} \cup \{\frac{1}{\mathcal{M}^s_\gamma(y)}\}$
    \ENDFOR
    \STATE $\bm{D} \gets \bm{D} \cup rand(6000, \bm{X}, norm(\bm{W}))$
\ENDFOR
\end{algorithmic}
\end{algorithm}

\subsection{Extended Experimental Settings}

\subsubsection{Implementations and Computational Resources}
We describe the implementation and compute details of the experiments in this section. For StyleGAN2 + ADA \cite{karras2020training} and StyleGAN2 \cite{karras2020analyzing}, we use the NVLabs codebase \footnote{\url{https://github.com/NVlabs/stylegan2-ada-pytorch}}. We employ the PyTorch-StudioGAN library \footnote{https://github.com/POSTECH-CVLab/PyTorch-StudioGAN} for ReACGAN + DiffAug \cite{kang2021rebooting, zhao2020differentiable}, BigGAN + DiffAug \cite{brock2018large, zhao2020differentiable}, BigGAN + CR \cite{brock2018large, Zhang2020Consistency}. We also adapt the official implementations for Projected GAN \cite{Sauer2021NEURIPS} \footnote{\url{https://github.com/autonomousvision/projected_gan}}, Improved DDPM \cite{song2020denoising, nichol2021improved} \footnote{\url{https://github.com/openai/improved-diffusion}}, NVAE \cite{vahdat2020NVAE} \footnote{\url{https://github.com/NVlabs/NVAE}}, and pseudo numerical methods for diffusion models (S-PNDM, F-PNDM, FON, PF, DDIM) \cite{song2020denoising, liu2022pseudo} \footnote{\url{https://github.com/liaopeiyuan/PNDM}}.

We mainly conduct our experiments with NVIDIA GPUs. See \tbl{appendix:comp} for the detailed configuration.

\begin{table}[h]
  \centering
  \resizebox{\textwidth}{!}{
  \begin{tabular}{llcc}
    \toprule
    Setting & Model     & GPU type     & Number of GPUs   \\
    \midrule
    $32 \times 32$, conditional & StyleGAN2 + ADA \cite{karras2020training} & TITAN RTX & 1   \\
    & StyleGAN2  \cite{karras2020analyzing} & TITAN RTX & 1  \\
    & ReACGAN + DiffAug \cite{kang2021rebooting, zhao2020differentiable} & RTX 2080Ti & 4  \\
    & Projected GAN \cite{Sauer2021NEURIPS} & RTX 2080Ti & 8  \\
    & BigGAN + DiffAug \cite{brock2018large, zhao2020differentiable} & RTX 2080Ti & 4  \\
    & BigGAN + CR \cite{brock2018large, Zhang2020Consistency} & RTX 2080Ti & 4 \\
    \midrule
    $32 \times 32$, unconditional & S-PNDM \cite{song2020denoising, liu2022pseudo} & RTX 2080Ti & 1 \\
    & F-PNDM \cite{song2020denoising, liu2022pseudo} & RTX 2080Ti & 1  \\
    & FON \cite{song2020denoising, liu2022pseudo} & RTX 2080Ti & 1  \\
    & PF \cite{song2020denoising, liu2022pseudo} & RTX 2080Ti & 1   \\
    & DDIM \cite{song2020denoising, liu2022pseudo} & RTX 2080Ti & 1   \\
    & Improved DDPM \cite{song2020denoising, nichol2021improved} & TITAN RTX & 1  \\
    & NVAE \cite{vahdat2020NVAE} & RTX 8000 & 4  \\
    & StyleGAN2 + ADA \cite{karras2020training} & TITAN RTX & 1  \\
    & StyleGAN2 \cite{karras2020analyzing} & TITAN RTX & 1  \\
    \midrule
    $256 \times 256$, conditional & Projected GAN \cite{Sauer2021NEURIPS} & RTX 8000 & 4 \\
    & StyleGAN2 + ADA \cite{karras2020training} & RTX 8000 & 4   \\
    & Latent Diffusion \cite{rombach2021highresolution} & RTX 8000 & 4   \\
    \midrule
    $256 \times 256$, impressionism, unconditional & Projected GAN \cite{Sauer2021NEURIPS} & RTX 2080Ti & 8 \\
    & Latent Diffusion \cite{rombach2021highresolution} & TITAN RTX & 4   \\
    \bottomrule
  \end{tabular}
  }
  \vspace{0.5em}
    \caption{Computational setup for models used in the benchmark. Across all models, only one node is used, and distributed training is done through data parallelism.}
     \label{appendix:comp}
\end{table}

\subsubsection{Hyperparameters}
We provide more details on hyperparameters used for each model across different experimental settings in \tbl{appendix:hparams}.

\begin{table}[h]
  \centering
  \resizebox{\textwidth}{!}{
  \begin{tabular}{llccc}
    \toprule
    Setting & Model     &  Training steps & Sampling steps & Batch size \\
    \midrule
    $32 \times 32$, conditional & StyleGAN2 + ADA \cite{karras2020training}& 390625 & \na & 64  \\
    & StyleGAN2  \cite{karras2020analyzing} & 117187 & \na & 64 \\
    & ReACGAN + DiffAug \cite{kang2021rebooting, zhao2020differentiable} &  200000 & \na & 128 \\
    & Projected GAN \cite{Sauer2021NEURIPS} & 156250 & \na & 64 \\
    & BigGAN + DiffAug \cite{brock2018large, zhao2020differentiable} & 200000 & \na & 128\\
    & BigGAN + CR \cite{brock2018large, Zhang2020Consistency}   & 200000 & \na & 128\\
    \midrule
    $32 \times 32$, unconditional & S-PNDM \cite{song2020denoising, liu2022pseudo} & 440000 & 250 & 128 \\
    & F-PNDM \cite{song2020denoising, liu2022pseudo}  & 440000 & 250 & 128 \\
    & FON \cite{song2020denoising, liu2022pseudo} & 440000 & 250 & 128 \\
    & PF \cite{song2020denoising, liu2022pseudo} & 440000 & 1000 & 128 \\
    & DDIM \cite{song2020denoising, liu2022pseudo} & 440000 & 1000 & 128 \\
    & Improved DDPM \cite{song2020denoising, nichol2021improved}  & 300000 & 1000 & 128 \\
    & NVAE \cite{vahdat2020NVAE} & 312500 & \na & 64 \\
    & StyleGAN2 + ADA \cite{karras2020training}  & 390625 & \na & 64  \\
    & StyleGAN2 \cite{karras2020analyzing}  & 116546 & \na & 64 \\
    \midrule
    $256 \times 256$, conditional & Projected GAN \cite{Sauer2021NEURIPS}  & 156250 & \na & 64 \\
    & StyleGAN2 + ADA \cite{karras2020training} & 390625 & \na & 64 \\
    & Latent Diffusion \cite{rombach2021highresolution} & 132812 & 250 & 64 \\
    \midrule
    $256 \times 256$, impressionism, unconditional & Projected GAN \cite{Sauer2021NEURIPS}  & 156250 & \na & 64 \\
    & Latent Diffusion \cite{rombach2021highresolution} & 93837 & 200 & 64 \\
    \bottomrule
  \end{tabular}
  }
  \vspace{0.2em}
   \caption{Hyperparameters for each model used in the benchmark, including number of training steps, number of diffusion sampling steps (if applicable), and batch size.}
   \label{appendix:hparams}
\end{table}

\subsection{Additional Details on Previous Artworks Datasets} \label{apppend:prev_datasets}
\myparagraph{VGG Paintings} \cite{Crowley2014TheSO} is a subset of the ‘Your Paintings’ dataset (\url{https://artuk.org/}) and contains 8,629 oil paintings of 10 classes (aeroplane, bird, boat, chair, cow, dining-table, dog, horse, sheep, train). The resolution of images is around 500 pixels in width. It was constructed by searching annotations and painting titles for corresponding classes in the PASCAL VOC dataset \cite{Everingham2009ThePV}.

\myparagraph{SemArt} \cite{Garca2018HowTR} is a dataset of 21,383 European fine-art paintings during the 8th-19th centuries collected from the Web Gallery of Art (\url{https://www.wga.hu/}). The associated CSV file contains images attributes, including Author, Title, Date, Technique, Type, School, and Timeframe. For preprocessing, images are scaled down to $256 \times 256$ resolution and randomly cropped into $224 \times 224$ resolution.

\myparagraph{WikiPaintings} \cite{Karayev2014RecognizingIS} is a dataset containing a total of 85,000 images collected from \url{Wikipaintings.org}. It has artworks (mostly paintings) labeled in 25 styles ranging from Renaissance to modern art movements and each style has more than 1000 images. 

\myparagraph{PrintArt} \cite{Carneiro2012ArtisticIC} is a dataset containing 988 monochromatic artistic images annotated by art historians with a global semantic annotation, a local compositional annotation, and a pose annotation of human subjects and animal types. It has 75 visual classes.

\myparagraph{BAM} \cite{Wilber2017BAMTB} is a large-scale dataset of contemporary artwork from Behance (\url{https://www.behance.net/}). This dataset can be found at \url{https://bam-dataset.org/}. It is consisted of 65 million images labeled by a human-in-the-loop automatic annotation pipeline. 3 types of annotations are provided, which are media attributes, emotion attributes, and object category attributes.

\myparagraph{Open MIC} \cite{Koniusz2018MuseumEI} is a dataset containing 16,156 images of 10 distinct source-target subsets, each from a different kind of museum exhibition spaces such as paintings, pottery and sculptures etc. designed for domain adaptation, egocentric recognition and few-shot learning. The dataset is segmented into 866 exhibit labels. Some images have low quality as they are collected via wearable camera, which suffered from sensor noises, motion blur, occlusions, background clutter, varying viewpoints etc.

\myparagraph{NoisyArt} \cite{DelChiaro2019NoisyArtAD} is a dataset designed for webly-supervised recognition of artworks, which contains more than 90,000 images in 3,120 webly-supervised classes. This dataset can be accessed at \url{https://github.com/delchiaro/NoisyArt}. It is collected from public knowledge bases like DBpedia at first, and then queried from Google Image Search and Flickr. It is a multi-modal and weakly-supervised dataset with noisy labels.

\myparagraph{Rijksmuseum} \cite{Mensink2014TheRC} is a dataset designed for visual classification and content-based retrieval of artistic content containing 112,039 artworks of ancient times, medieval ages and the late 1900s exhibited in the Rijksmuseum Amsterdam, retrieved online from \url{https://www.rijksmuseum.nl/en/research/conduct-research/data}. All images are stored at 300 dpi. Attributes created for different tasks, which are artist, art-type, material, and creation year predictions.

\myparagraph{iMET} \cite{Zhang2019TheIC} is a dataset for fine-grained artwork attribute recognition, which contains 155, 531 images of 1103 attributes collected from The Met museum. Each image is resized so that the shorter side is 300 pixels and stored in PNG format. The dataset is annotated with culture-related attributes from Subject Matter Experts (SME) derived attributes and vendor-sourced attributes provided by The Met.

\myparagraph{The Met} \cite{ypsilantis2021met} is a dataset for large-scale instance-level recognition of artworks consisted of 418k exhibit images corresponding to about 224k unique exhibits, where each exhibit is a class. All images are resized to have maximum resolution $500 \times 500$. Labels are created from the corresponding Met classes, and a rigorous filtering, annotation, and verification labeling process is employed to annotate samples without query metadata.

\myparagraph{Art500k} \cite{Mao2017DeepArtLJ} is a large-scale visual arts dataset of over 500,000 digital artworks scraped from scraped from WikiArt (\url{https://www.wikiart.org/}), Web Gallery of Art (\url{http://www.wga.hu/}), Rijks Museum (\url{https://www.rijksmuseum.nl/nl}) and Google Arts \& Culture (\url{https://www.google.com/}). It is publicly available at \url{http://deepart2.ece.ust.hk/ART500K/art500k.html}. Images are annotated with corresponding Wiki pages information or results of searching titles via Google. MD5 hashing is used to remove duplicates from the scraped data. The dataset has multiple categories, which are Artist, Genre, Medium, History Figure and Event, and each category contains its own classes.

\myparagraph{OmniArt} \cite{Strezoski2018OmniArtAL} is an artworks dataset consisted of over 2 million images with rich structured metadata, which contains 1,348,017 indexed images with full annotations and 702,000 unlabeled images. It provides various types of attributes through concepts, IconClass labels, color information, and object-level bounding boxes. The dataset has both meta-level metadata (artwork genre, subtype, color palettes etc.) and object-level metadata (bounding boxes), and can be used for benchmarking a wide range of tasks including artist attribution, creation period estimation, style prediction and so on.

\subsection{Generated Samples from Model (256 $\times$ 256)}
\begin{figure}[H]
\centering
\includegraphics[width=\linewidth]{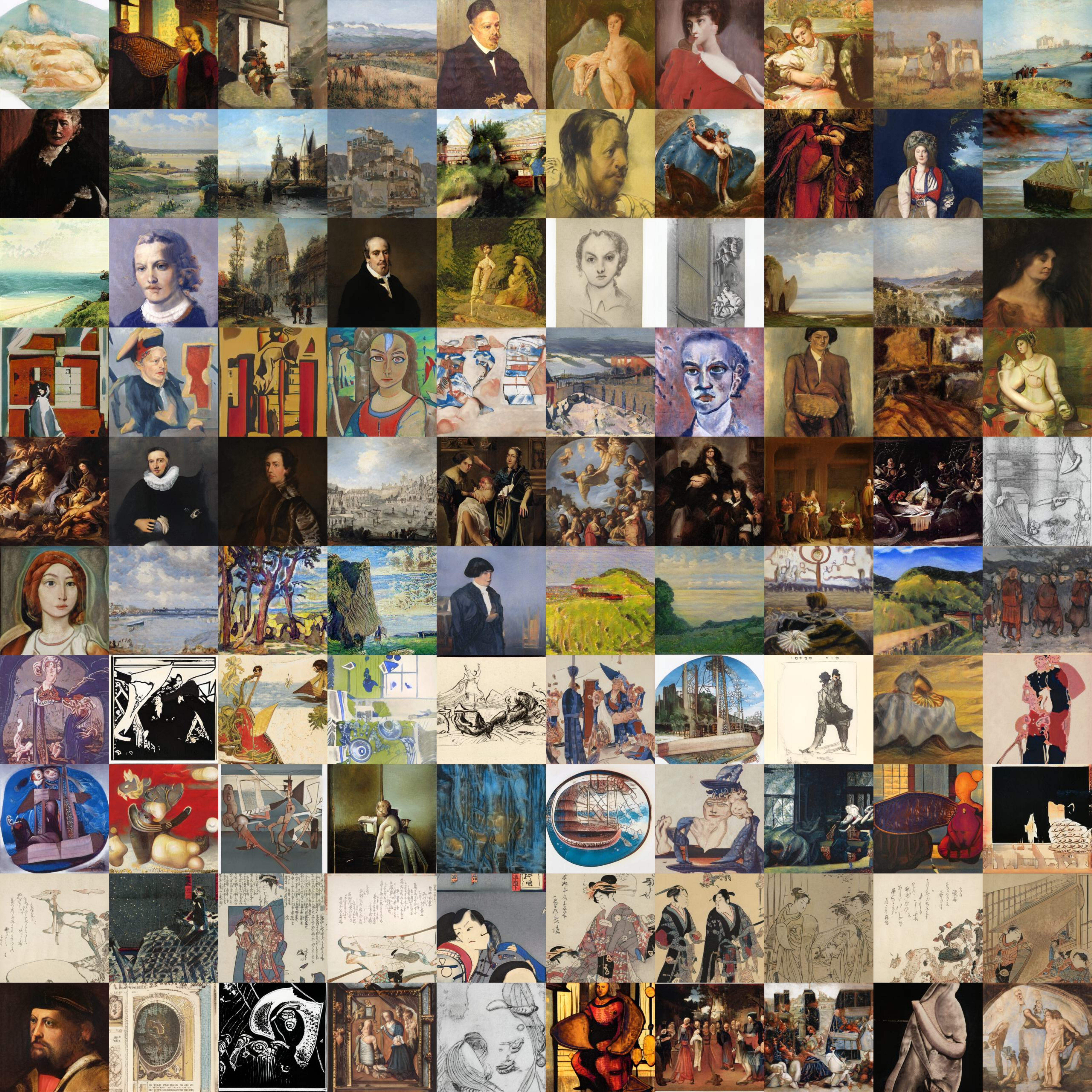}
\caption{Samples from our best conditional $256\times256$ model, StyleGAN2 + ADA \cite{karras2020training}. The images are selected randomly given one global random seed. Images in each row come from a single class, which are impressionism, realism, romanticism, expressionism, baroque, post impressionism, art nouveau, surrealism, ukiyo-e, renaissance from top to bottom.}
\vspace{-5pt}
\label{append:samples_256}
\end{figure}

\newpage

\subsection{Generated Samples from Model (32 $\times$ 32)}
\begin{figure}[H]
\centering
\includegraphics[width=\linewidth]{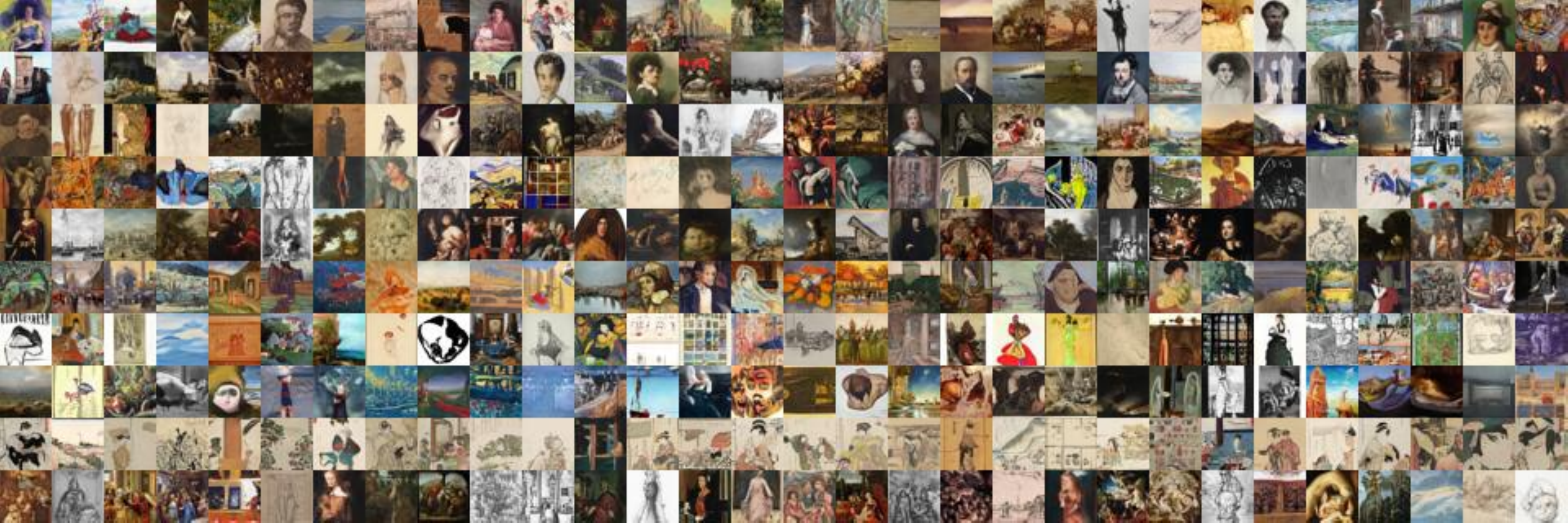}
\caption{Samples from our best conditional $32\times32$ model, StyleGAN2 + ADA \cite{karras2020training}. The images are selected randomly given one global random seed.  Images in each row come from a single class, which are impressionism, realism, romanticism, expressionism, baroque, post impressionism, art nouveau, surrealism, ukiyo-e, renaissance from top to bottom.}
\vspace{-5pt}
\label{append:samples_32}
\end{figure}

\newpage

\subsection{k-NN Retrieval for Generated Samples (256 $\times$ 256)}
\begin{figure}[H]
\centering
\includegraphics[width=\linewidth]{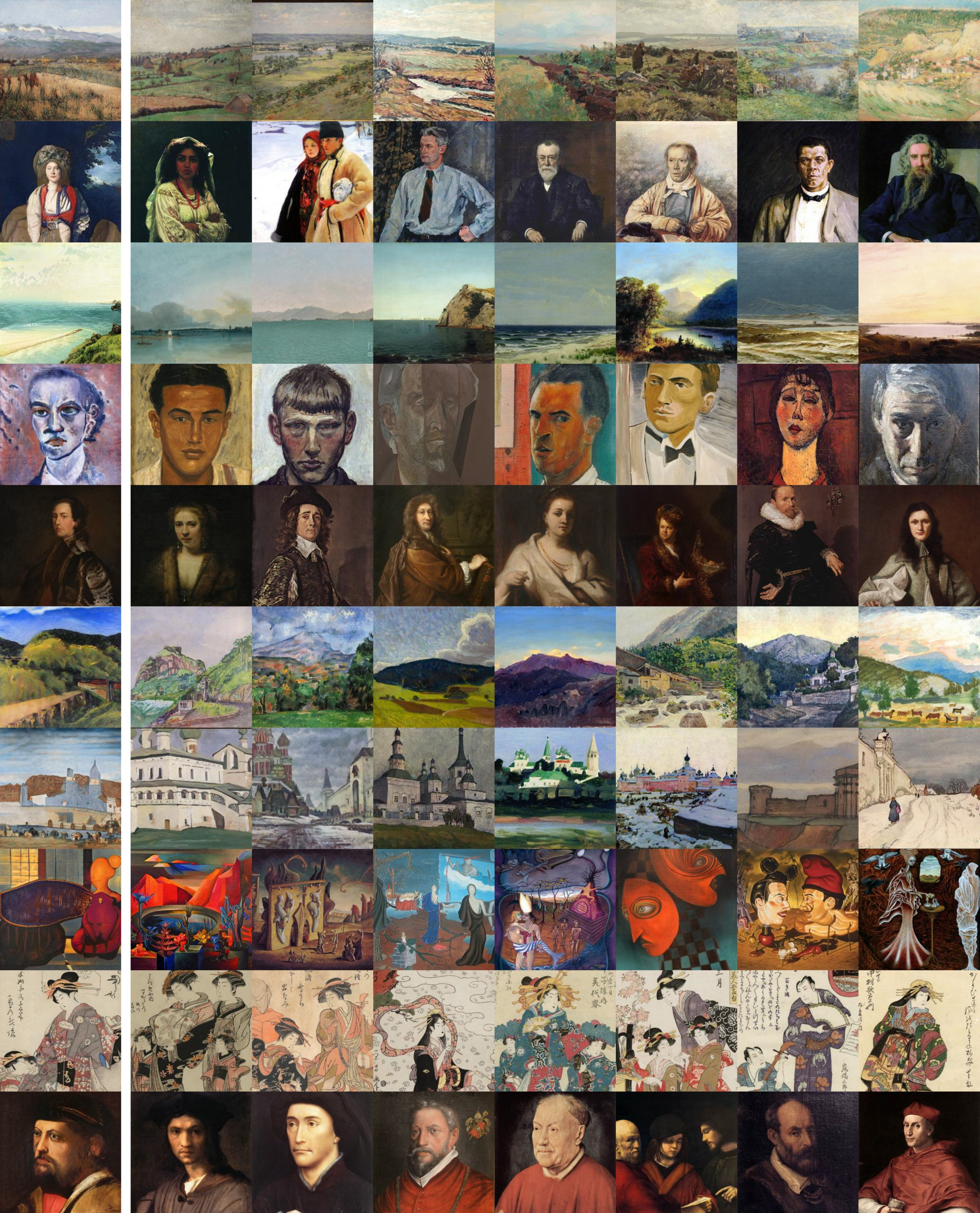}
\caption{Top-7 nearest neighbors retrieval results for selected samples generated from our StyleGAN2 + ADA \cite{karras2020training} ($256 \times 256$) model. The first column is generated images as queries, and the other columns are retrieved images from training set. Images in each row come from a single class, which are impressionism, realism, romanticism, expressionism, baroque, post impressionism, art nouveau, surrealism, ukiyo-e, renaissance from top to bottom.}
\vspace{-5pt}
\label{append:knn}
\end{figure}

\end{document}